\documentclass{naturep}

\usepackage{amssymb}
\usepackage{amsmath}
\usepackage{graphicx}
\usepackage{algorithmicx}
\usepackage{algorithm}

\usepackage[noend]{algpseudocode}
\usepackage{bbm}
\usepackage{lineno}
\usepackage[margin=0.6in]{geometry}
\usepackage{ragged2e}
\usepackage{setspace}
\usepackage{longtable}
\usepackage{hyperref}
\usepackage{anyfontsize}
\usepackage{setspace}
\usepackage{float}
\usepackage{booktabs}
\usepackage{multirow}
\usepackage{xcolor}
\usepackage{blindtext}
\usepackage{tabularx}
\usepackage{caption}
\usepackage{subcaption}

\makeatletter
\let\saved@includegraphics\includegraphics
\AtBeginDocument{\let\includegraphics\saved@includegraphics}

\makeatother

\usepackage{enumitem, kantlipsum}

\title{\begin{flushleft}{\begin{spacing}{1}Towards a Visual-Language Foundation Model for Computational Pathology\end{spacing}}\end{flushleft}}

\begin{document}

\maketitle
\begin{spacing}{1.8}
\vspace{-15mm}
\noindent Ming Y. Lu$^{1,2,3,4,6\boldsymbol{\ddag}}$, Bowen Chen$^{1,2\boldsymbol{\ddag}}$, Drew F. K. Williamson$^{1,2,3\boldsymbol{\ddag}}$, Richard J. Chen$^{1,2,3,4,5}$, Ivy Liang$^{1,8}$, Tong Ding$^{1}$, Guillaume Jaume$^{1,2,3,4}$, Igor Odintsov$^{1}$, Andrew Zhang$^{1,2,3,4,7}$, Long Phi Le$^{2,7}$, Georg Gerber$^{1}$, Anil V Parwani$^{9}$, Faisal Mahmood$^{*1,2,3,4,5,10}$
\end{spacing}
\vspace{-6mm}
\begin{spacing}{1.4}
\begin{affiliations}
 \item Department of Pathology, Brigham and Women's Hospital, Harvard Medical School, Boston, MA
 \item Department of Pathology, Massachusetts General Hospital, Harvard Medical School, Boston, MA
 \item Cancer Program, Broad Institute of Harvard and MIT, Cambridge, MA 
 \item Cancer Data Science Program, Dana-Farber Cancer Institute, Boston, MA
 \item Department of Biomedical Informatics, Harvard Medical School, Boston, MA
 \item Electrical Engineering and Computer Science, Massachusetts Institute of Technology (MIT), Cambridge, MA
 \item Health Sciences and Technology, Harvard-MIT, Cambridge, MA
 \item Harvard John A. Paulson School of Engineering And Applied Sciences, Harvard University, Cambridge, MA
 \item Department of Pathology, Wexner Medical Center, Ohio State University, Columbus, OH
 \item Harvard Data Science Initiative, Harvard University, Cambridge, MA
 \\$\boldsymbol{\ddag}$ Contributed Equally
 \\\textbf{*Corresponding author}: Faisal Mahmood (faisalmahmood@bwh.harvard.edu)
\end{affiliations}
\end{spacing}


\begin{spacing}{1.2}
\noindent \textbf{The accelerated adoption of digital pathology and advances in deep learning have enabled the development of powerful models for various pathology tasks across a diverse array of diseases and patient cohorts\cite{huang2022deep, litjens20181399, campanella2019clinical, coudray2018classification, lu2021data, skrede2020deep, porpoise, courtiol2019deep, lu2021ai, zhu2023accurate, sish, yottixel, smily}. However, model training is often difficult due to label scarcity in the medical domain and the model's usage is limited by the specific task and disease for which it is trained\cite{huang2022deep, litjens20181399, campanella2019clinical,bulten2020automated,nagpal2019development}. Additionally, most models in histopathology leverage only image data, a stark contrast to how humans teach each other and reason about histopathologic entities. We introduce CONtrastive learning from Captions for Histopathology (CONCH), a visual-language foundation model developed using diverse sources of histopathology images, biomedical text, and notably over 1.17 million image-caption pairs via task-agnostic pretraining. Evaluated on a suite of 13 diverse benchmarks, CONCH can be transferred to a wide range of downstream tasks involving either or both histopathology images and text, achieving state-of-the-art performance on histology image classification, segmentation, captioning, text-to-image and image-to-text retrieval. CONCH represents a substantial leap over concurrent visual-language pretrained systems for histopathology, with the potential to directly facilitate a wide array of machine learning-based workflows requiring minimal or no further supervised fine-tuning.}
\end{spacing}

\newpage

\begin{spacing}{1.35}
\noindent\textbf{\large{Introduction}} 

The gold standard for the diagnosis of many diseases remains the examination of tissue by a pathologist. The recent rise of computational pathology (CPath)\cite{abels2019computational}, which leverages artificial intelligence (AI) to solve problems in pathology, has demonstrated considerable advances across many tasks including metastasis detection\cite{huang2022deep, litjens20181399, campanella2019clinical, bejnordi2017diagnostic}, cancer subtyping\cite{coudray2018classification, lu2021data}, survival prediction\cite{skrede2020deep, porpoise, lee2022derivation, courtiol2019deep, lu2020prognostic}, unknown primary origin site prediction\cite{lu2021ai, zhu2023accurate}, image search\cite{sish, yottixel, smily}, and mutation prediction\cite{kather2020pan, saldanha2023self}, among other tasks\cite{echle2021deep,beck2011systematic,yala2022optimizing,tarantino2021evolving,madabhushi2016image, zhou2023multi, laleh2022benchmarking, shmatko2022artificial, graham2019hover}.
Additionally, current strides in the field are made under the paradigm of developing models targeting specific tasks using large cohorts of labeled training examples, such as in lymph node metastasis detection\cite{huang2022deep, litjens20181399, campanella2019clinical,bejnordi2017diagnostic} and prostate cancer grading\cite{bulten2020automated, nagpal2019development}. However, the process of data collection and annotation of whole slide images (WSIs) is labor-intensive and is not scalable to open-set recognition problems or rare diseases, both of which are common to the practice of pathology. With thousands of possible diagnoses and many other tasks, training separate models for every step of the pathology workflow is untenable. Additionally, as diverse as these tasks are, they are all analyses of visual data or include other structured information such as ``omics``\cite{mobadersany2018predicting,cheerla2019deep, chen2021multimodal,wang2023shared,kather2020pan,fu2020pan, carrillo2023rna,jing2020multi,sammut2022multi,boehm2022multimodal} and other multimodal data sources~\cite{weng2019multimodal, dwivedi2022multi,huang2023artificial, foersch2023multistain,vanguri2022multimodal}. However, the practice of pathology and the communication of pathological findings make extensive use of natural language, be it in the form of the report that the pathologist prepares for the patient and their treating clinician, the journal article that details a new histopathologic entity, or the textbook chapter that teaches residents how to practice pathology.

The general machine learning community has made immense strides in foundation models that utilize both visual and language information. Representative works such as CLIP\cite{radford2021learning}, ALIGN\cite{jia2021scaling}, and CoCa\cite{yu2022coca}, among others\cite{li2022blip, schuhmann2022laion, dou2022empirical, li2021align, chen2020uniter,singh2022flava,li2023uni,lu2019vilbert,alayrac2022flamingo,li2023scaling,wang2023image,recasens2023zorro} use large-scale image-caption pairs to pretrain visual-language foundation models---task-agnostic pretrained models that demonstrate robust performance in downstream vision and visual-language tasks.    
In the broader biomedical imaging domain, visual-language data have been leveraged for a variety of tasks including X-ray report generation\cite{boag2020baselines,endo2021retrieval,chen-etal-2020-generating,miura-etal-2021-improving,pmlr-v106-liu19a}, zero-shot classification\cite{tiu2022expert,huang2021gloria,zhang2022contrastive, zhang2023large, wang2022medclip}, retrieval\cite{schaumberg2020interpretable, maleki2022lile, huang2021gloria, zhang2022contrastive, wang2022medclip}, among others\cite{zhang2023pathnarratives,tsuneki2022inference,zhang2020evaluating,naseem2022vision,he2021towards}. 
However, the number of works integrating vision and language data for representation learning in CPath is small, with recent works\cite{huang2023leveraging, zhang2023large, gamper2021multiple, lu2023visual, lin2023pmc} demonstrating the potential of using paired image-caption data to learn meaningful visual representations and to develop foundation models for histopathology that can transfer to multiple downstream tasks in a zero-shot setting, \textit{i.e.}, using no task-specific training data. However, these works\cite{zhang2023large, lu2023visual, huang2023leveraging} are limited in the scale of histopathology-specific pretraining data due to the lack of readily-available image-caption pairs in this domain, leading to limited practical utility from relatively poor performance. Additionally, the broader capabilities of these models remain underexplored. 

Given the diversity of tasks, the difficulty in acquiring large datasets of rare diseases or combinations of findings, and the central nature of language to the practice of pathology, there is a need for 1) high-performing visual-language foundation models that leverage large-scale pretraining and generalize well across tasks and 2) extensive study on the wide range of potential applications of these models in order to understand their utilities and limitations. We introduce CONtrastive learning from Captions for Histopathology (CONCH), a visual-language foundation model developed using diverse sources of histopathology images, biomedical text, and over 1.17 million image caption pairs (\textbf{Figure 1a-c, Extended Data Figure 1}) via task-agnostic pretraining with the aim to address these unfilled needs. Based on CoCa\cite{yu2022coca}, a state-of-the-art visual-language foundation pretraining framework, CONCH utilizes an image encoder, a text encoder, and a multimodal fusion decoder, and is trained via a combination of contrastive alignment objectives that seek to align the image and text modalities in the model's representation space and a captioning objective that learns to predict the caption corresponding to an image (\textbf{Figure 1d}). We investigate the capabilities of CONCH on a wide array of tasks, including classification of image tiles and gigapixel WSIs, cross-modal image-to-text and text-to-image retrieval, image segmentation, and image captioning in a total of thirteen diverse benchmarks. We demonstrate that our model achieves state-of-the-art performance across all benchmarks relative to other visual-language foundation models (\textbf{Figure 1e}), including PLIP\cite{huang2023leveraging}, BiomedCLIP\cite{zhang2023large}, OpenAICLIP\cite{radford2021learning} and outperforms concurrent baselines often by a large margin (\textbf{Figures 2-6}).

\begin{figure*}
\centering
\includegraphics[width=\textwidth]{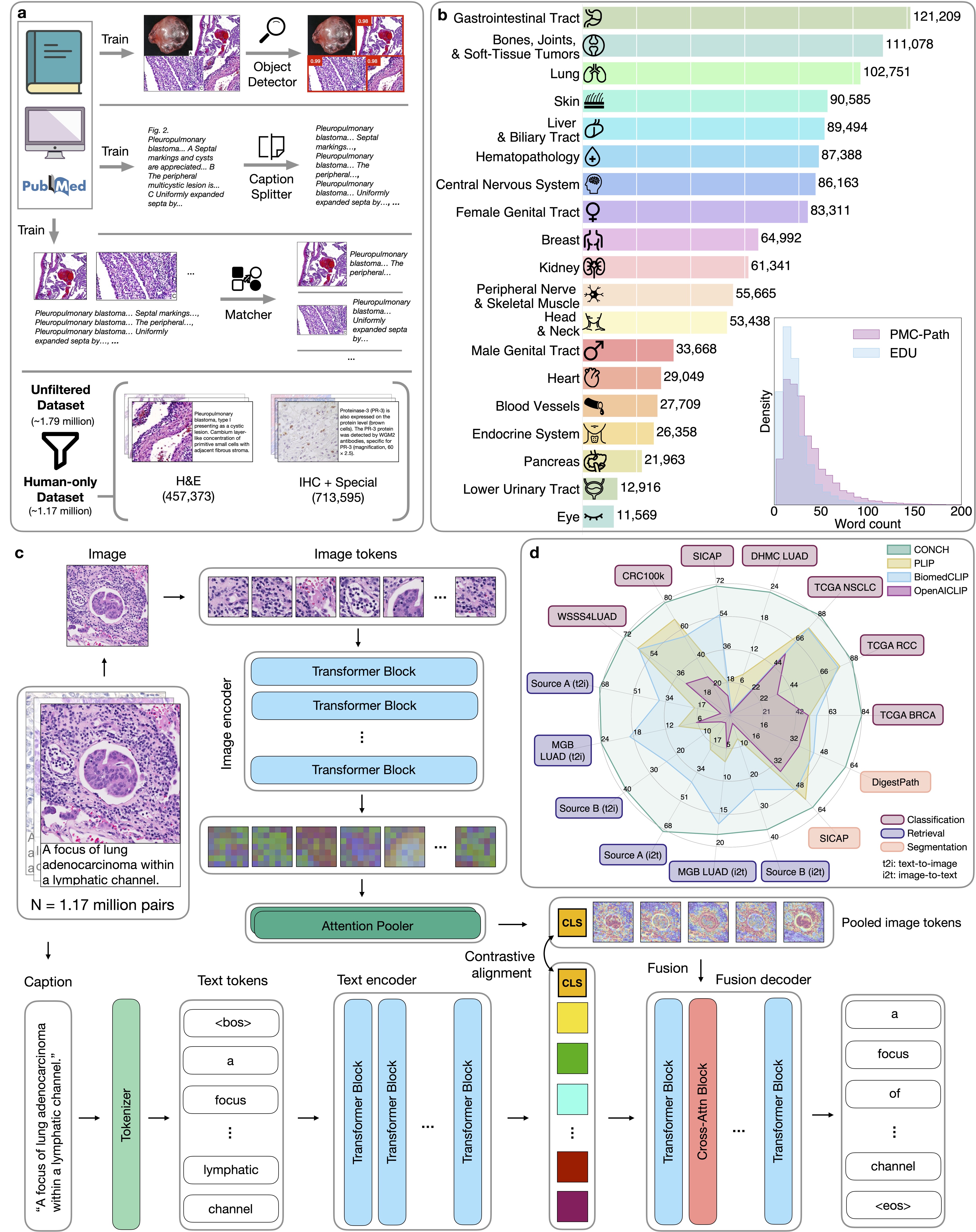}
\caption*{\textbf{Figure 1: Data curation and model schematic.} Caption on next page.}
\end{figure*}
\begin{figure*}
  \caption*{(Previous page.) \textbf{Figure 1: Data curation and model schematic.} \textbf{a}. Automated data cleaning pipeline. Educational sources (EDU) and parts of the PubMed Central Open Access Dataset (PMC OA) were manually cleaned and used to train an object detector to detect histopathology images, a language model to split captions referring to multiple images, and a matching model to match detected images to their corresponding captions. The cleaning process yields a dataset of 1.79 million image-text pairs, which we then filter out pairs referring to non-humans to create our CONCH (human-only) pretraining dataset of 1.17 million. See \textbf{Methods} for details on data cleaning and \textbf{Extended Data Figure 9} on performance comparisons using different variations of the pretraining dataset. \textbf{b.} Estimated distribution of image-text pairs in the human-only pretraining dataset by topic. Note that pretraining data covers a diverse range of pathology topics. Inset compares distribution of caption lengths between PMC-Path and EDU. See \textbf{Extended Data Figure 1} for wordclouds of captions from each category. \textbf{c.} Visual-language pretraining setup. CONCH consists of an image encoder, a text encoder, and a multimodal text decoder. The pretraining process uses both contrastive and captioning objectives. The contrastive objectives align the image and text encoders by maximizing the cosine-similarity scores between paired image and text embeddings while the captioning objective maximizes the likelihood of generating the correct text conditioned on the image and previously generated text. See \textbf{Methods} for details. \textbf{d.} Radarplot comparing performance of CONCH and baselines on various downstream tasks. CONCH outperforms baselines by a significant margin on a diverse set of tasks spanning classification, retrieval, and segmentation. See \textbf{Results} for detailed descriptions of each task and metrics.}
\end{figure*}

\noindent\textbf{\large{Results}}

\noindent\textbf{Zero-shot classification of diverse tissues and diseases}\\
Contrastively-aligned visual-language pretraining allows the model to be directly applied to downstream classification tasks without requiring further labeled examples for supervised learning or finetuning. This zero-shot transfer capability allows a single pretrained foundation model to be applied off-the-shelf to many different downstream datasets with arbitrarily many classes compared with the current paradigm of training a new model for every new task. Given a task, we first represent the set of class or category names using a set of predetermined text prompts where each prompt corresponds to a class. An image is then classified by matching it with the most similar text prompt in the model's shared image-text representation space (\textbf{Figure 2a}, see \textbf{Methods} for details). In practice, there are often multiple ways to phrase the same concept in text (\textit{e.g.} ``invasive lobular carcinoma of the breast" and ``breast ILC") so we ensemble multiple text prompts for each class during prediction, which was found to generally boost predictive performance compared to using a single text prompt (\textbf{Extended Data Figure 2}). Additionally, while previous works\cite{zhang2023large, huang2023leveraging} primarily focused on classification tasks at the region-of-interest (ROI) level, we also investigate the zero-shot capability of our model on gigapixel whole slide images (WSIs) by leveraging MI-Zero\cite{lu2023visual}, which divides a WSI into smaller tiles and subsequently aggregates individual tile-level scores into a slide-level prediction (\textbf{Figure 2b}). 

In total, we evaluate on four slide-level classification tasks: TCGA BRCA (invasive breast carcinoma subtyping), TCGA NSCLC (non-small cell lung cancer subtyping), TCGA RCC (renal cell carcinoma subtyping), and DHMC LUAD (lung adenocarcinoma histologic pattern classification) and three ROI-level tasks: CRC100k (colorectal cancer tissue classification), WSSS4LUAD (lung adenocarcinoma tissue classification), and SICAP (Gleason pattern classification). We use balanced accuracy as the primary evaluation metric for TCGA NSCLC, TCGA RCC, TCGA LUAD, CRC100k and WSSS4LUAD, which accounts for class imbalance by weighing the accuracy score of each class equally. Following the community standard, we use Cohen's $\kappa$ and quadratic weighted Cohen's $\kappa$ as primary metrics for lung adenocarcinoma (LUAD) pattern classification and Gleason pattern classification respectively, as they are regarded as more subjective tasks, which typically translates to higher inter-rater variability. We refer readers to \textbf{Extended Data Tables 1-14} for more detailed reporting of model performance and \textbf{Methods} for detailed descriptions of evaluation datasets.

On slide-level benchmarks, CONCH outperforms state-of-the-art visual-language foundation models (PLIP, BiomedCLIP, and OpenAICLIP) on all tasks, often by a wide margin (\textbf{Figure 2c}). For instance, for non-small cell lung cancer (NSCLC) subtyping and renal cell carcinoma (RCC) subtyping, CONCH achieves a zero-shot accuracy of 90.0\% and 89.3\% respectively, and outperforms the next best performing model, PLIP, by 11.3\% and 8.9\% on each task with $p < 0.01$ via two-sided paired permutation test (see \textbf{Statistical analysis}). On the more difficult invasive breast carcinoma (BRCA) subtyping task, CONCH achieves a zero-shot accuracy of 84.0\% while other models perform near random chance at between 50.7\% (PLIP) to 55.3\% (BiomedCLIP), nearly 30\% ($p < 0.01$) lower than CONCH. Similarly, on the more challenging LUAD pattern classification task, CONCH achieved a $\kappa$ score of 0.236, almost 0.16 higher than the next highest performing model, PLIP ($p = 0.014$). On ROI-level benchmarks, we observe similar findings, where CONCH achieves a zero-shot quadratic $\kappa$ of 0.711 on SICAP (outperforming BiomedCLIP by 0.158, $p < 0.01$), a zero-shot accuracy of 79.1\% on CRC100k (outperforming PLIP by 11.7\%, $p < 0.01$) and a zero-shot accuracy of 71.9\% on WSSS4LUAD (outperforming PLIP by 9.5\%, $p < 0.01$). These results demonstrate that in addition to achieving more accurate predictions on relatively easy tasks, CONCH is still able to achieve meaningful predictions on more challenging tasks where other models may potentially struggle to outperform random chance.
\begin{figure*}
\centering
\vspace{-9mm}
\includegraphics[width=0.95\textwidth]{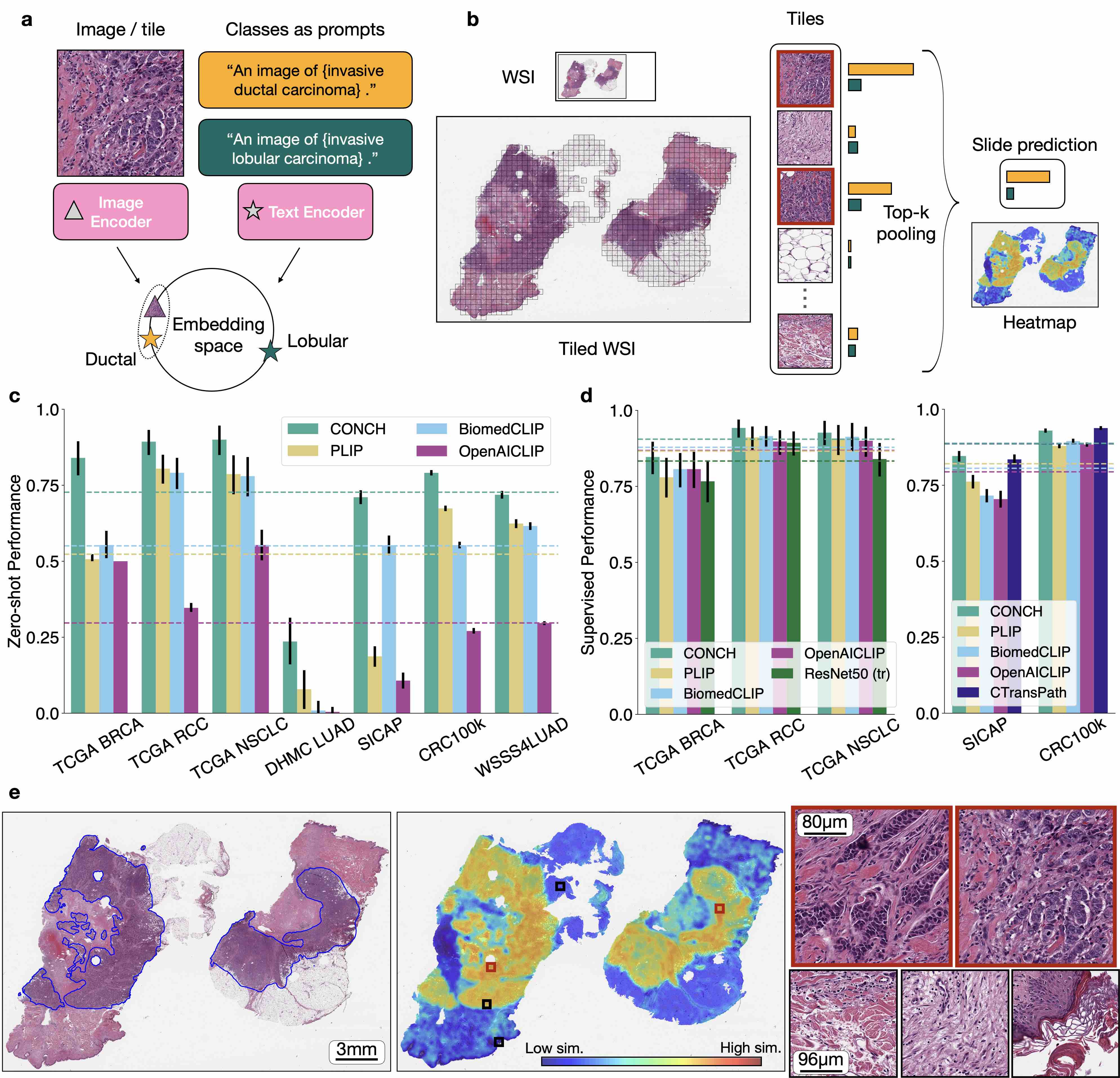}
\caption*{\textbf{Figure 2: Zero-shot and supervised classification.} \textbf{a.} Schematic of zero-shot classification using a pair of contrastively aligned image and text encoders. A prompt is constructed for each class, and the image is classified according to the prompt whose embedding is closest to that of the image in the shared embedding space. \textbf{b.} Zero-shot classification of WSIs. Each WSI is divided into tiles and processed as in \textbf{a}. The similarity scores for tiles are aggregated using top-$K$ pooling to form slide-level similarity scores, the highest of which corresponds to the slide-level prediction. In \textbf{c, d}, dashed lines represent the average over tasks. Error bars represent 95\% confidence intervals. \textbf{c.} Zero-shot performance on downstream subtyping (TCGA BRCA, $n=150$; TCGA RCC, $n=225$; TCGA NSCLC, $n=150$; DHMC LUAD, $n=143$; CRC100k, $n=7,180$; WSSS4LUAD, $n=4,693$) and grading (SICAP, $n=2,122$) tasks. Cohen's $\kappa$ is reported for DHMC LUAD and quadratically weighted Cohen's $\kappa$ is reported for SICAP, while balanced accuracy is reported for all other tasks. Additional metrics are reported in \textbf{Extended Data Tables 1-7}. \textbf{d.} Supervised evaluation of embeddings of each model. Linear probing is used for ROI-level tasks (CRC100k and SICAP) while ABMIL is used for slide-level tasks, with the same metrics reported as in \textbf{c.}. See \textbf{Extended Data Tables 15-19} for more detailed results. \textbf{e.} From left to right: pathologist-annotated invasive ductal carcinoma (IDC), corresponding heatmap, and selected tiles at higher power. Heatmap is colored based on cosine-similarity score between each tile within the slide and the text prompt corresponding the predicted class label. We find excellent agreement between the annotated image and high-similarity regions, with the tiles demonstrating classic IDC morphology within the high-similarity regions and stroma or other normal constituents of the breast in the low similarity regions.}
\end{figure*}

When classifying a WSI using zero-shot transfer, in addition to computing an aggregated, slide-level prediction, we can also create a heatmap to visualize the cosine-similarity score between each tile in the slide and the text prompt corresponding the predicted class label. Regions with high similarity scores are deemed by the model to be close matches with the diagnosis (\textit{e.g.} invasive ductal carcinoma, IDC) while regions with low similarity scores do not match the diagnosis (\textbf{Figure 2e}). In an example of breast IDC slide, we find that regions highlighted in the heatmap closely resemble the tumor regions as delineated by pathologist annotation (\textbf{Figure 2e, left} and \textbf{middle}). Since the slide-level prediction score is a simple average of the similarity scores of the top-K tiles for a given class, the heatmap enables human interpretability by directly highlighting regions involved in the model's decision making process, which can be displayed in high resolution to the human user for inspection (\textbf{Figure 2e, center}). Additional examples are visualized in \textbf{Extended Data Figure 3-5}. These findings suggest the possibility of using the zero-shot recognition ability of our model for coarse-grained tissue segmentation on WSIs, which we quantitatively evaluate in the \textbf{Zero-shot segmentation} section.

\noindent\textbf{Few-shot classification with task-specific supervised learning}\\
The zero-shot recognition capability of contrastive pretrained visual-language models for histopathology enables efficient and expedited application of a single foundation model to a potentially wide range of tasks without going through the laborious processes of training data collection, annotation, and supervised model training for each new task. Sometimes however, it may still be desirable to specialize the model with labeled training examples in order to maximize performance for a given task, ideally using as few labels as possible. In this section, we investigate the label efficiency in using the pretrained representation of the image encoder backbone of the visual-language foundation models for task-specific supervised classification. For each benchmark using supervised training, we either use the official training set (if provided) or remaining cases from the dataset after holding out the set of cases used for zero-shot evaluation (see \textbf{Downstream evaluation datasets}). For slide-level tasks, we train weakly-supervised classification models using slide-level labels based on the widely used Attention-based multiple instance learning algorithm\cite{ilse2018attention}. For ROI-level tasks, we use logistic regression on top of the global (\textit{e.g.} $\textless$\texttt{CLS}$\textgreater$ token) representation of each encoder, a practice commonly known as linear probing. In addition to PLIP, BiomedCLIP, and OpenAICLIP encoders, we add additional baselines for comparison: for slide-level tasks, given its popularity, ResNet50\cite{he2016deep} (truncated after the third residual block) pretrained on ImageNet\cite{deng2009imagenet}; and for ROI-level tasks, CTransPath\cite{wang2022transformer}---a state of the art self-supervised pretrained histopathology image encoder (see \textbf{Methods} for details).

On the slide-level tasks (\textbf{Figure 2d, left}), CONCH achieves a balanced accuracy score of 84.7\%, 94.2\% and 92.7\% on BRCA subtyping, RCC subtyping and NSCLC subtyping respectively, outperforming the commonly used ResNet50 ImageNet baseline by 8.0\%, 4.9\% and 8.7\% respectively ($p = 0.027$,  $p < 0.01$ and $p = 0.033$). Overall, CONCH obtained an average accuracy of 90.5 \% across the three tasks vs. PLIP and BiomedCLIP at 86.6\% and 87.9\% respectively, but no statistical significance was detected for each task. In the ROI-level tasks (\textbf{Figure 2d, right}), CONCH performs nearly identically as the state-of-the-art CTransPath encoder (93.0\% vs. 93.8\% balanced accuracy on CRC100k and 0.846 vs. 0.835 quadratic weighted $\kappa$ on SICAP), while outperforming PLIP, BiomedCLIP and OpenAICLIP by between by 3.4\% and 5.1\% in balanced accuracy on CRC100k and between 0.084 and 0.142 in quadratic weighted $\kappa$ on SICAP ($p < 0.01$ for all comparisons). These results demonstrate that overall, CONCH provides a strong image encoder that either performs comparably to or better than all visual encoders tested, including the self-supervised state-of-the-art method. See \textbf{Extended Data Tables 15-19} for detailed reporting of model performance. 

\begin{figure*}
\centering
\includegraphics[width=\textwidth]{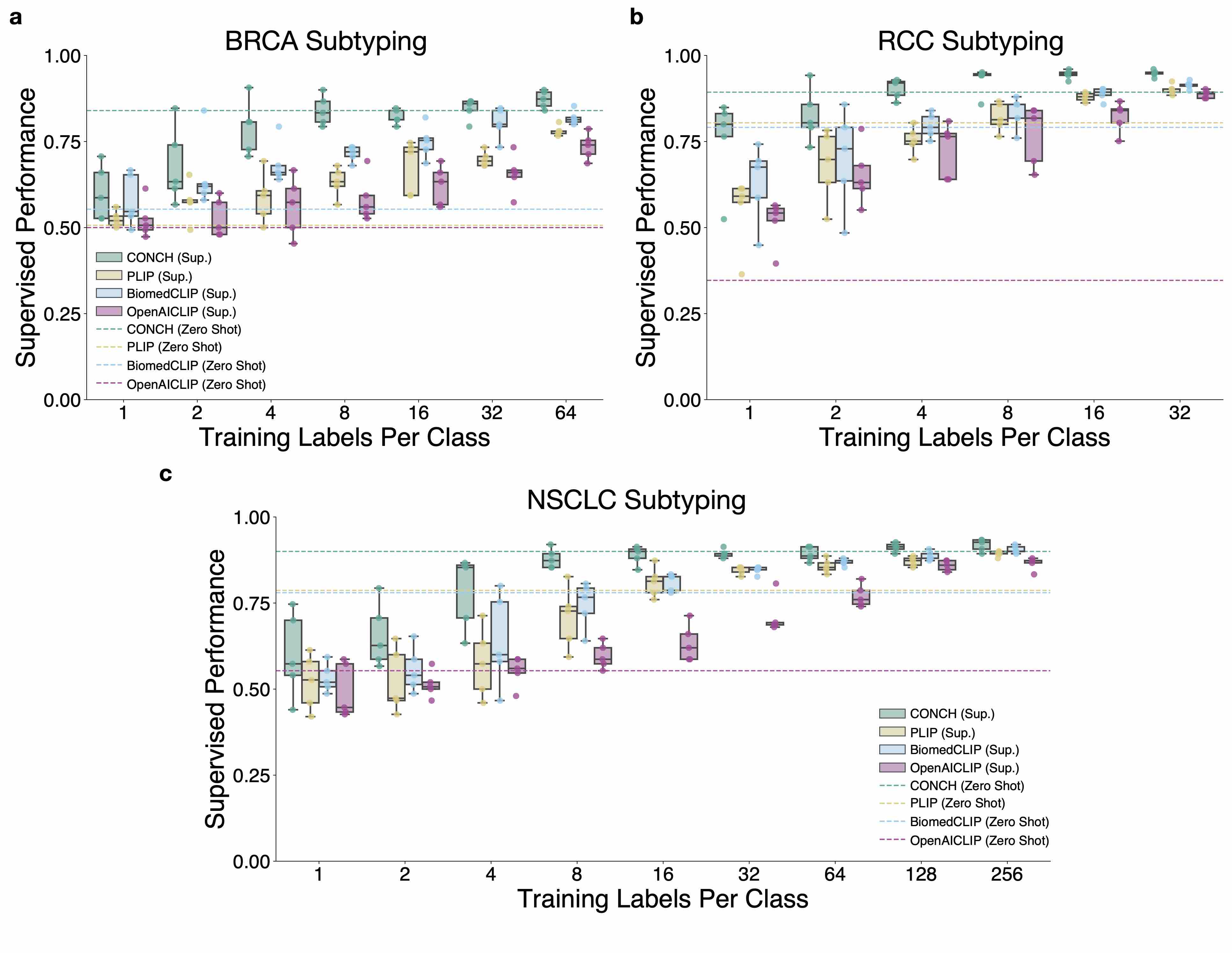}
\caption*{\textbf{Figure 3: Slide-level few-shot classification experiments.} We investigate the label efficiency of different visual-language pretrained encoders in the few-shot setting where we vary the number of training labels per class ($n_{c}$), for $n_{c} = 1, 2, 4, 8, 16 \ldots$ until we reach the maximum number of available labels in the training set. For each $n_c$, we sample 5 different sets of training examples and train a weakly-supervised ABMIL model on each training set using slide-level labels (see \textbf{Supervised classification experiments} for details). We show their individual model performance via boxplot (\textit{i.e.}, $n=5$ for each box) to study the variance in model performance when performing supervised learning with very few training examples. Boxes indicate quartile values and whiskers extend to data points within 1.5$\times$ the interquartile range. For reference, the zero-shot performance of each model is shown as a dotted line on the same plot. In terms of few-shot supervised learning, CONCH achieves better performance (\textit{i.e.} in terms of the median accuracy of 5 runs) than other encoders for different sizes of training set and for all tasks. Additionally, CONCH zero-shot performance is surprisingly competitive, outperforming PLIP, BiomedCLIP, and OpenAICLIP few-shot up to 64 labels per class in the case of BRCA and NSCLC subtyping. 
}
\end{figure*}

Next we investigate the label efficiency of different visual-language pretrained encoders in the few-shot setting where we vary the number of training labels per class ($n_{c}$), for $n_{c} = 1, 2, 4, 8 \ldots$ up to 512 per class or until we reach the maximum number of available labels in the training set. In the few shot setting, for each experiment, we sample 5 different sets of training examples and show their individual performance via boxplot to account for the high variance in model performance when performing supervised learning with very few training examples (\textbf{Figure 3} and \textbf{Extended Data Figure 6}). We first observe that CONCH achieves better performance (in terms of median accuracy of 5 runs) than other encoders for all sizes of training set and for all tasks, which translates to requiring fewer labels to achieve the same performance. 
For instance in BRCA subtyping, using the CONCH encoder and 8 training labels per class outperforms using PLIP, BiomedCLIP or OpenAICLIP with 64 labels per class, representing an 8$\times$ reduction in training set size---a trend we also observe for most tasks tested. Additionally, we note that the zero-shot performance of CONCH is highly competitive when compared to few-shot supervised learning. Aside from relatively easy tasks such as RCC subtyping and CRC tissue classification, CONCH zero-shot in fact outperforms PLIP and BiomedCLIP-based supervised learning in BRCA subtyping (up to 64 labels per class), NSCLC subtyping (up to 128 labels per class), and Gleason grading (up to 16 labels per class for PLIP and 64 labels per class for BiomedCLIP). These findings suggest that the zero-shot capability of a good visual-language foundation model should not be trivialized and in fact can serve as a very good baseline when evaluating the performance of task-specific diagnostic models trained with supervised learning. On the other hand, the zero-shot capability of previous visual-language foundation models (\textit{i.e.} PLIP and BiomedCLIP) can be easily surpassed by the CONCH encoder and supervised learning using just a few examples (1 label per class for BRCA subtyping and CRC tissue classification, 4 for RCC subtyping, NSCLC subtyping, and Gleason grading). 


\begin{figure*}
\centering
\vspace{-9mm}
\includegraphics[width=\textwidth]{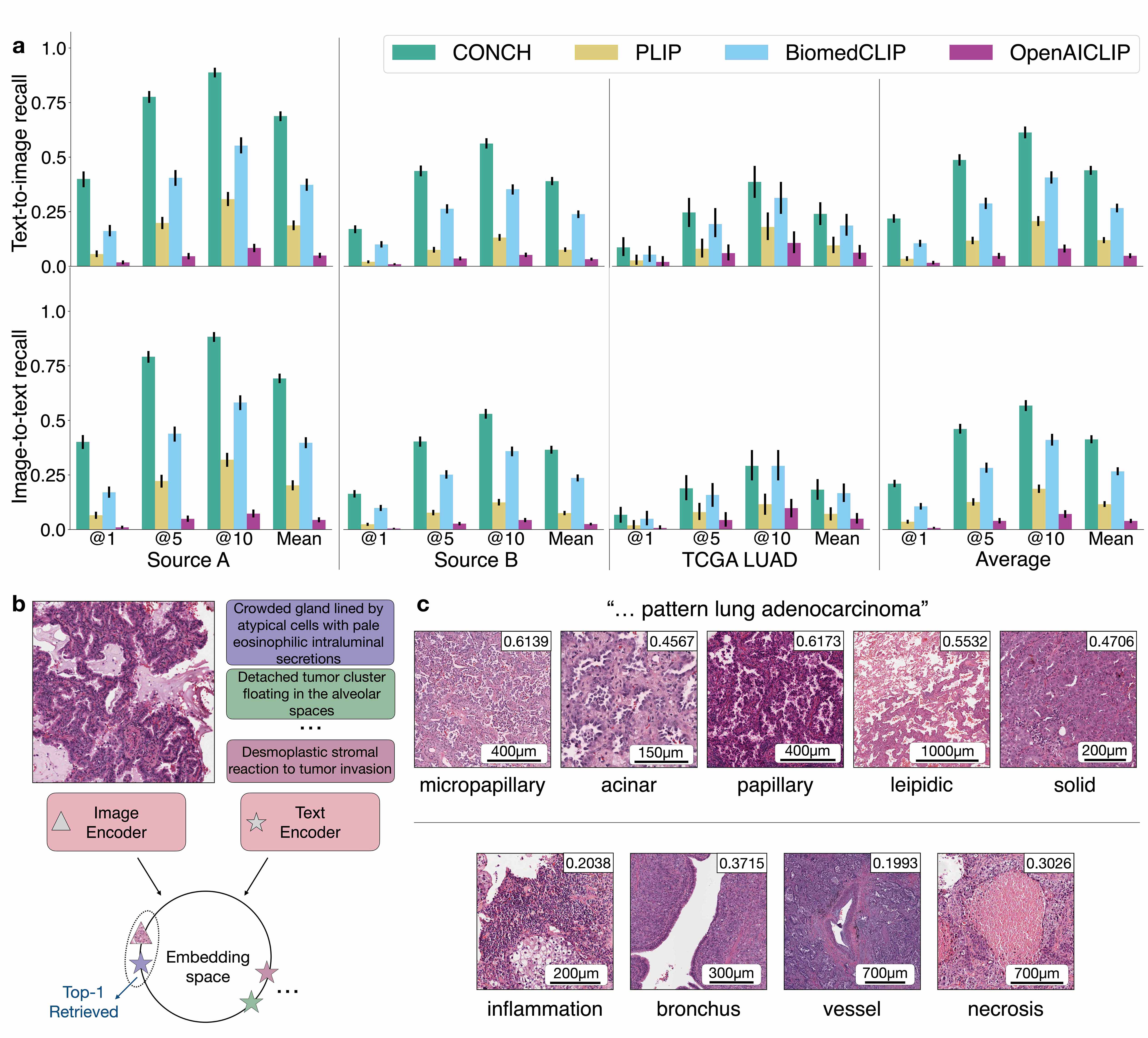}
\caption*{\textbf{Figure 4: Zero-shot Cross-Modal Retrieval.} \textbf{a.} Model performance in cross-modal retrieval was evaluated on 3 datasets of image-text pairs (Source A, $n=797$; Source B, $n=1,755$; TCGA-LUAD, $n=1,65$). Similarity in the embedding space is computed between the query image with all text samples in the database. The top-$K$ most similar texts are retrieved. 
We report Recall@$K$ for $K\in\{1,5,10\}$ as well as the Mean Recall, which averages over $K$. We show both text-to-image (\textbf{top} row) and image-to-text (\textbf{bottom} row) retrieval for each retrieval task (\textbf{columns}). The rightmost column reports the average across tasks for each metric. CONCH outperforms other baselines on all retrieval tasks. Error bars indicate 95\% confidence intervals. \textbf{b.} Schematic for zero-shot image-to-text retrieval (text-to-image is analogous). \textbf{c.} Examples of images in top-5 retrieved results from TCGA LUAD using LUAD-relevant queries with cosine-similarity scores shown in top-right corner. Examples for other datasets using more diverse queries are shown in \textbf{Extended Data Figure 7}. In general, we find the images retrieved by the model match what is described in the text prompt.}
\end{figure*}

\noindent\textbf{Zero-shot cross-modal retrieval}\\
By learning an aligned latent space for visual and language embeddings, our model is capable of cross-modal retrieval in a zero-shot setting, \textit{i.e.}, retrieving the corresponding text entry based on an image query (image-to-text, abbreviated as ``i2t"), or vice versa (text-to-image, abbreviated as ``t2i"). This task naturally lends itself to image search applications, which are useful in the biomedical domain for applications such as identifying cases for inclusion in research cohorts or clinical trials, assistance with rare disease presentations or morphologies, and collecting cases for or helping to create educational resources. To perform text-to-image retrieval (the image-to-text direction is analogous), we use the text encoder to embed a text input that serves as a query. We then use the query text embedding to retrieve similar images in the latent space (\textbf{Figure 4b)}. 

We evaluate our model on three image-caption datasets, Source A and Source B (both are held-out sources from scraping that cover a diverse range of general pathology concepts) and TCGA LUAD (a much more specific dataset of tiles extracted from LUAD slides in TCGA and annotated with captions in house). Following previous works\cite{jia2021scaling,huang2023leveraging,zhang2023large}, we use Recall@$K$ as the metric for cross-modal retrieval. See the \textbf{Methods} section for more detailed descriptions of retrieval datasets.

On average over the three datasets, CONCH significantly outperforms baselines by a large margin, achieving mean recall for text-to-image retrieval of 44.0\% and outperforms the next best model, BiomedCLIP, by 17.3\% with $p<0.01$ via two-sided paired permutation test (Figure 4a). For Source A and Source B, CONCH achieves mean recall for text-to-image retrieval of 68.8\% and 39.0\% respectively, outperforming the second highest model, BiomedCLIP, by 31.5\% and 15.1\% ($p<0.01$ for both). For TCGA LUAD, CONCH achieves text-to-image mean recall of 24.0\%, outperforming the next best model, BiomedCLIP, by 5.3\% but with no statistical significance ($p=0.22$). However, CONCH outperforms PLIP and OpenAICLIP significantly ($p<0.01$). Image-to-text retrieval for all three datasets follows the same trend as text-to-image retrieval in terms of performance and statistical significance, except for TCGA LUAD where the gap for CONCH and BiomedCLIP is slightly smaller (1.6\%). We refer readers to \textbf{Extended Data Tables 20-25} for more detailed reporting of model performance. Based on these results, CONCH is able to perform more accurate cross-modal retrieval compared to baselines. 

Aside from using the paired captions as queries, we also show examples of retrieved results using CONCH with simple text prompts of concepts related to LUAD (\textit{e.g.}, ``solid pattern lung adenocarcinoma") on the TCGA LUAD dataset (\textbf{Figure 4c}) and general pathology concepts (\textit{e.g.}, ``melanoma") on Source A and Source B (\textbf{Extended Data Figure 7a-b}). To provide examples from more complex text queries, such as ``cribriform prostatic adenocarcinoma" or ``IDH wildtype glioma", we used a highly diverse dataset of 321,261 tiles sampled from 1,620 cases held-out during pretraining, spanning 108 OncoTree\cite{kundra2021oncotree} codes (\textbf{Extended Data Figure 7c}). However, as this dataset does not have paired text data, we were not able to quantify the retrieval performance. The presented examples are confirmed by a pathologist to represent the text query closely.

\begin{figure*}
\centering
\includegraphics[width=\textwidth]{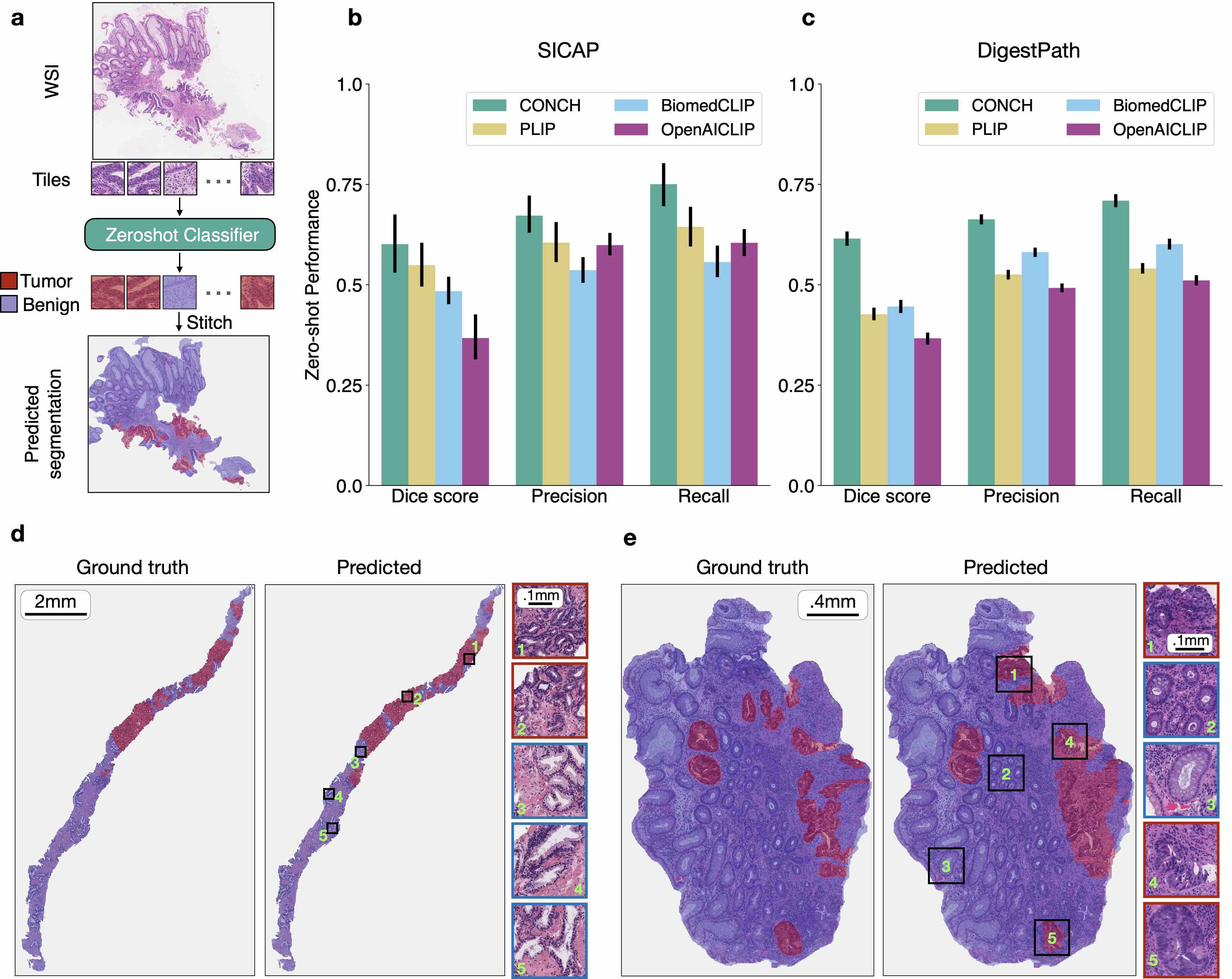}
\caption*{\textbf{Figure 5: Zero-shot Segmentation.} \textbf{a.} Schematic illustrating zero-shot segmentation on WSIs (or large tissue sections). To perform segmentation, we divide each WSI into tiles and use zero-shot classification to predict the label of each tile. The tile-level predictions are stitched together to form the predicted segmentation mask. \textbf{b-c.} Zero-shot segmentation performance of CONCH and baselines on SICAP ($n=31$) and DigestPath ($n=250$) dataset respectively. The macro-averaged Dice score, precision and recall are reported. Error bars represent 95\% confidence intervals. \textbf{d-e.} Example of CONCH segmentation prediction on WSIs. \textbf{Left} panel shows ground truth and \textbf{right} panel shows predicted segmentation mask, with example regions enlarged. Red and blue indicate tumor and normal tissue respectively. In general, in these examples, CONCH displays excellent sensitivity to tumor regions with slightly lower specificity, though most of the regions that CONCH segments as tumor which are in fact non-tumor are adjacent to cancerous glands or contain cancer-associated stroma for both SICAP and DigestPath.}
\end{figure*}

\noindent\textbf{Zero-shot segmentation}\\
While WSIs can be gigapixels in size, they are generally heterogeneous, with diverse cell types, morphologies, and tissue architectures represented, each often comprising a small share of the slide. Consequently, segmentation on the slide level is a difficult and useful task to identify distinct regions of a WSI based on characteristics of interest and can reduce the number of tiles needed for downstream applications. However, since annotated data at the sub-slide level is expensive and laborious to collect, a general model capable of performing slide-level segmentation in a zero-shot setting is valuable. In this work, we explore the possibility of performing coarse-grained tissue segmentation on WSIs without labeled examples but instead directly using the aforementioned demonstrated zero-shot retrieval and classification capabilities of our model.

Given a WSI, we divide the tissue regions into into smaller image tiles and pose a given segmentation task as classifying each tile using zero-shot classification, and assigning the predicted class label to all pixels in the tile, performed for all tiles. To minimize sharp transition in predicted values for pixels at the boundary of neighboring tiles, we tile the WSIs with a 75\% overlap and average the prediction scores in overlapped regions in order to achieve a smoother appearance in the predicted segmentation map. We evaluate our model on SICAP for prostate tumor vs. normal tissue segmentation and on DigestPath for malignant vs. benign tissue segmentation in colorectal cancer specimens. We report dice score, precision, and recall for each task against ground truth pixel-level annotations, with scores macro-averaged over all images in each dataset (see \textbf{Methods} for more details). We refer the reader to \textbf{Extended Data Tables 26-27} for more detailed results of model performance. 

CONCH outperforms other models in both tasks (\textbf{Figure 5a, c}). In SICAP, CONCH achieves a average dice score of 0.601 (PLIP: 0.549, $p = 0.08$ and BiomedCLIP: 0.484, $p < 0.01$), an average recall score of 0.751 (PLIP: 0.644, $p < 0.01$, BiomedCLIP: 0.557, $p < 0.01$), and an average precision core of 0.672 (PLIP: 0.605, $p = 0.024$, BiomedCLIP: 0.536, $p < 0.01$). In DigestPath, CONCH achieves a average dice score of 0.569 (PLIP: 0.374, $p < 0.01$ and BiomedCLIP: 0.408, $p < 0.01$), an average recall score of 0.684 (PLIP: 0.513, $p < 0.01$, BiomedCLIP: 0.576, $p < 0.01$), and an average precision core of 0.644 (PLIP: 0.495, $p = 0.024$, BiomedCLIP: 0.559, $p < 0.01$). Additionally, we find that despite the coarse-grained and zero-shot nature of the approach, the model was able to produce reasonably accurate pixel-level segmentation masks in some instances, as visualized in \textbf{Figure 5b, d}.

\begin{figure*}
\centering
\includegraphics[width=\textwidth]{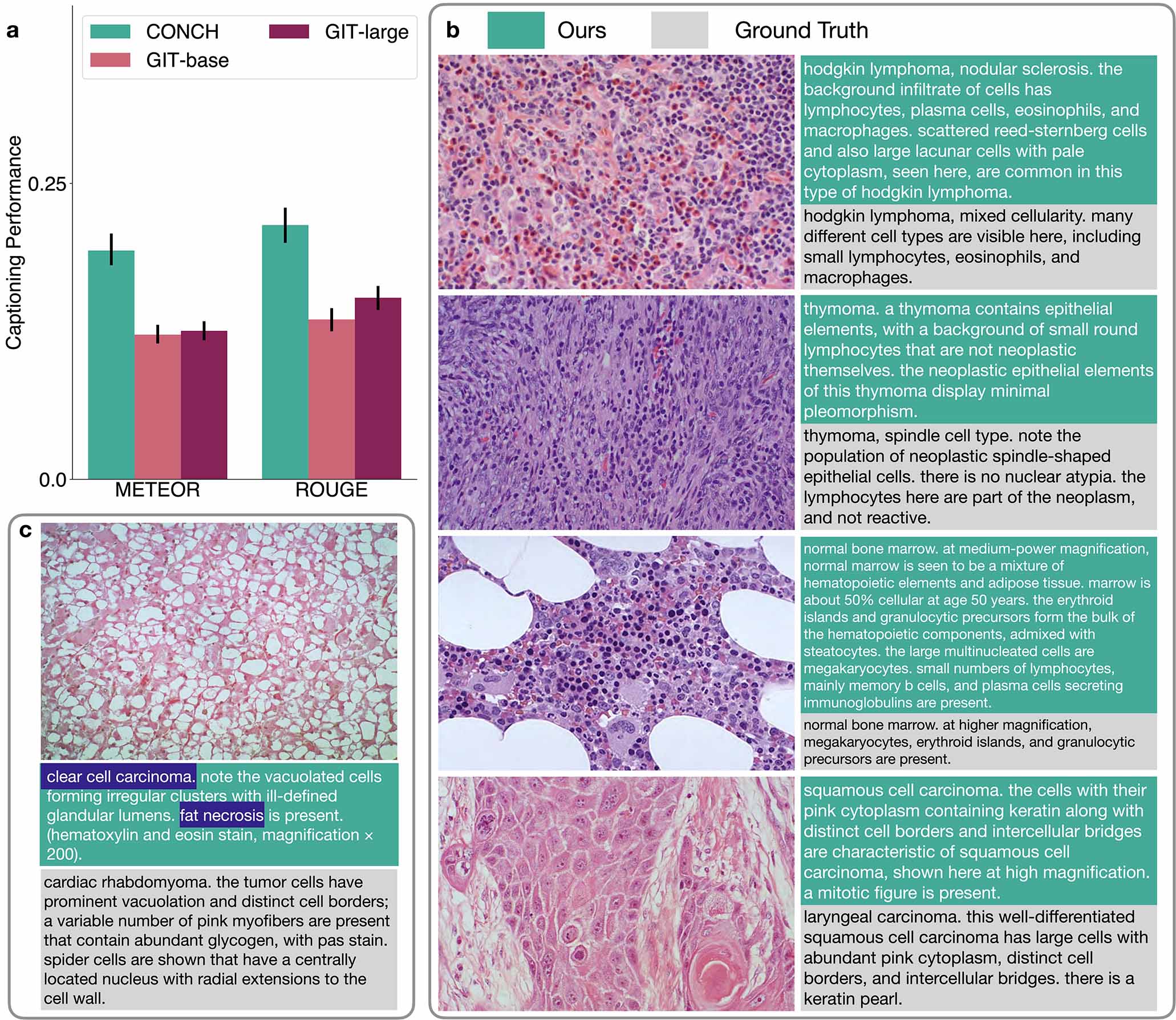}
\caption*{\textbf{Figure 6: Captioning.} \textbf{a.} Captioning performance of CONCH and baselines fine-tuned on Source A (train $n=558$, validation $n=77$, test $n=162$). The METEOR and ROUGE metrics are both calculated to evaluate the quality of generated captions. Captions were generated using top-$K$ sampling with $K=50$ as the decoding strategy. Error bars represent 95\% confidence intervals. \textbf{b.} Examples of captions generated by CONCH considered by a pathologist to be high quality. The green text boxes show generated captions and gray text boxes show ground truth captions. \textbf{c.} An example of an incorrect caption generated by CONCH. Though the diagnosis and description is incorrect, the image does resemble clear cell renal cell carcinoma or adipose tissue with fat necrosis at low resolution upon review by a pathologist, indicating that CONCH is able to recognize some features shared by these disparate diseases (incorrect but reasonable text highlighted in blue).}
\end{figure*}

\noindent\textbf{Captioning}\\
Image captioning has been a widely explored task in the general visual-language domain\cite{wang2022git,li2023blip,alayrac2022flamingo}. On top of distilling a top-level diagnosis of the image, image captioning can potentially provide morphological and contextual details as well as additional interpretability, offering a much richer set of information than discrete labels. While prior works\cite{lu2023visual,huang2023leveraging,zhang2023large} in visual-language pretraining have shown applications in classification and retrieval, they are not equipped with generative capabilities. By adding a generative loss along with alignment and a text encoder module using the CoCa framework, our model is augmented with the ability to generate text conditioned on image inputs. We explore the captioning capabilities of CONCH, laying down the first work to explore image captioning using a vision language foundation model in the histopathology domain. 

For this task, we use image-caption pairs extracted from a held-out source, Source A, where a board certified pathologist manually reviewed and condensed each caption such that it retains only information that can be inferred from the image, including the top-level diagnosis and detailed morphological descriptions. Given that our pretraining data is far from the scale of high quality zero-shot captioning, we perform finetuning on the dataset. We partition the dataset into training, validation, and testing splits and fine-tune CONCH and baselines. We refer the reader to \textbf{Methods} for more details on the dataset and finetuning. Since PLIP and BiomedCLIP are not readily adaptable to captioning tasks, we compare against a GenerativeImage2Text (GIT)\cite{wang2022git}, a widely-used family of open-source visual language pretrained models for image captioning. To measure the quality of generated captions, we report METEOR and ROUGE, two widely used metrics used for image captioning. 

On the captioning task (\textbf{Figure 6a}), CONCH achieves a METEOR score of 0.193 and a ROUGE score of 0.215, outperforming all baselines (GIT-base: METEOR 0.122, ROUGE 0.135 and GIT-large: METEOR 0.125, ROUGE 0.153) with $p < 0.01$. Although our absolute performance on these metrics is not ideal, image captioning is a considerably more difficult task than classification and retrieval, and we show that our pretraining data and approach can significantly improve performance over general visual-language models. While we find that our model is able to generate captions that strongly relate to the contents of image inputs (See \textbf{Figure 6b-c} for examples), we noticed that some of the generated captions are regurgitated verbatim from the training dataset, likely due to the limited scale of fine-tuning (training split $n=558$). Given that our current pretraining scale is still relatively small compared to works in the general visual-language domain, we expect the fine-tuned captioning performance to improve and potentially even achieve quality zero-shot captioning. Our work makes one of the first strides in this underexplored direction in histopathology. We refer the reader to \textbf{Extended Data Table 28} for more detailed reporting of model performance. 


\noindent\textbf{\large{Discussion}} \\
Most previous works in computational pathology have attempted to extract meaningful patterns and discriminative signals from image data and/or structured patient data such as genomics and have ignored the textual aspect of pathology. However, these approaches leave on the table a huge amount of information present in descriptions of images, information that allows pathology trainees to generalize from a few exemplar images of an entity to images in the real world that are often significantly more diverse. While recent works have attempted to leverage image and caption data from social media or biomedical research articles to build visual-language foundation models applicable to the domain of histopathology, we found across a number of tasks that both their zero-shot and supervised classification performance remain limited, hindering their practical value as general purpose recognition or retrieval systems for histpathology. Additionally, beyond working on small ROIs, the models' abilities to perform in more complex settings (\textit{e.g.} classification or tumor segmentation on heterogeneous gigapixel WSIs) remain underexplored.

In this study, we demonstrated that by using the current largest histopathology-specific, paired image-text dataset of over 1.17 million examples for task-agnostic pretraining, we can build a high-performance visual-language foundation model that can then demonstrate utility in a wide range of clinically-relevant downstream tasks from classification, retrieval, segmentation to captioning. Our model is equipped with strong zero-shot recognition capabilities out of the box, which can potentially relieve the burden of annotating training examples for many specific classification tasks, as we demonstrated that its zero-shot performance often rivals or even outperforms conventional supervised learning baselines in these tasks under few-shot settings. Additionally, the much improved zero-shot image-to-text and text-to-image retrieval capabilities of our model will potentially empower trainees, physicians, and researchers to more accurately and flexibly retrieve relevant patient cases or educational examples based on image or natural language queries once it can be efficiently implemented into healthcare systems or databases. Equipped with a multimodal decoder, our visual-language foundation model also provides the flexibility to be further fine-tuned in downstream tasks that involve language generation (\textit{e.g.} image captioning) and/or multimodal reasoning based on both visual and textual inputs.

A key limitation of our study is the scale of data pretraining, which still pales when compared to billion-scale datasets used in developing large scale visual-language foundation models in the general machine learning community, and therefore we are likely to see further potential improvement in zero-shot recognition capabilities, representation quality, and robustness by increasing both the quantity and quality of histopathology image-caption datasets. Additionally, while the current landscape of visual-language foundation models for histopathology focuses primarily on image-level tasks, the ability of these models to recognize fine-grained visual concepts at the region-level (\textit{i.e.} cellular or even sub-cellular level) has not yet been studied, meaning that other important tasks such as mitosis detection, fine-grained tissue segmentation, or cell counting currently remain outside the scope of their downstream capabilities.





\noindent\textbf{\large{Online Methods}}

\noindent\textbf{Pretraining dataset curation}\\
We use publicly available articles from PubMed and educational resources to curate the largest-to-date dataset of histopathology image-caption pairs. We use deep learning to automate data cleaning iteratively. For curation, we divide the data sources into two categories: \textbf{EDU}, which consists of data extracted from educational sources, and \textbf{PMC OA}, which consists of data downloaded from the PubMed Central Open Access Dataset\footnote{\href{https://www.ncbi.nlm.nih.gov/pmc/tools/openftlist/}{ncbi.nlm.nih.gov/pmc/tools/openftlist/}}. 

The data curation process poses two main challenges: filtering for histopathology data and handling image panels. The first challenge is that the raw downloaded data comprise  both histopathology and non-histopathology examples. The second challenge is that a significant portion of EDU and most of PMC OA are in the form of figure panels, where the image consists of multiple sub-images arranged in a panel with parts of the caption addressing all or some of the sub-images. In light of these challenges, manually cleaning the data is infeasible. We clean the data in three steps: 1) detecting histopathology images (as single images or sub-images), 2) splitting captions that refer to image panels into separate captions into sub-captions, and 3) aligning sub-images with sub-captions within each image panel. To automate the cleaning process using deep learning, we take advantage of the fact that EDU is significantly cleaner than PMC OA and orders of magnitude smaller (45k \textit{vs.} 18M) by manually cleaning EDU and using it as the starting training data for each step described below. 


To detect histopathology images, we use an object detection model (YOLOv5)\cite{redmon2016you} to generate bounding boxes for extracting detected images. To avoid the laborious task of manually labeling ground truth bounding boxes in EDU, we generate synthetic data by randomly selecting single-panel images and arranging them in an image panel. We iteratively refine the detection model by validating on a small subset ($<0.5\%$) of PMC OA and adding incorrectly labeled samples to the training set. 

For caption splitting, we collected a dataset of original and split captions (while cleaning EDU) to fine-tune a GPT-style model pretrained on PubMed and other medical text\cite{luo2022biogpt}. We pose the problem of splitting captions as causal language modeling, where we fine-tune the language model to take the original full caption as input, and predict the sub-captions separated by the key word ``Next caption: ". We use the fine-tuned model to perform caption splitting. 

To align the detected histopathology images with split captions, we first train a CLIP model\cite{radford2021learning} on the cleaned EDU dataset along with PMC OA single figures that do not require splitting and alignment. Using the trained model, given a set of $m$ detected images and $n$ split captions from an image panel, we compute the image embeddings $\{\mathbf{u}_0, \mathbf{u}_1, \dots, \mathbf{u}_m\}$ and text embeddings $\{\mathbf{v}_0, \mathbf{v}_1, \dots, \mathbf{v}_n\}$ in the aligned latent space. For each image embedding $\mathbf{u}_i$, we compute the cosine-similarity score with each text embedding $\mathbf{v}_j$. We retrieve the text that has the highest cosine-similarity score $s_{i,j}:={\mathbf{u}_i}^T\mathbf{v}_j$ and consider $\{\mathbf{u}_i, \mathbf{v}_j\}$ to be an image-caption pair for our cleaned dataset.

By applying the three steps above to PMC OA, we create PMC-Path, a pathology-specific image-caption dataset derived from PubMed figures. We then combine it with EDU to form our full unfiltered pretraining dataset of 1,786,362 image-caption pairs. However, PMC-Path also contains a significant number of pairs that refer to animal histopathology as well as non-H\&E stains (such as IHC, Masson's trichrome, Congo red, \textit{etc.}). Since our downstream evaluation only concerns human histopathology and H\&E tasks, we would like to assess how the animal and special staining data would affect performance. We first parse the captions to exclude samples referencing non-human animals, forming a dataset of 1,170,647 human pairs. Additionally, we trained a classifier that identifies H\&E stains to further filter the human-only dataset and create a dataset of 457,372 pairs. We find that CONCH pretrained on the human-only dataset performed the best on downstream tasks in general (See \textbf{Extended Data Figure 8}). 

\noindent\textbf{Visual-language pretraining} \\
For visual-language pretraining, we use an equal-weighted combination of the image-text contrastive loss and the captioning loss following CoCa\cite{yu2022coca}, a state-of-the-art visual-language foundation model pretrained on general domain image-caption pairs. The model consists of an image encoder, $f(\,\cdot\,; \theta)$, a text encoder, $g(\,\cdot\,; \phi)$, and a multimodal text decoder, $h(\,\cdot\,; \psi)$. The image encoder includes the backbone and two attentional pooler modules, parameterized by $\theta_{\mathrm{backbone}}, \theta_{\mathrm{contrast}}$ and $\theta_{\mathrm{caption}}$ respectively. The backbone is a vision transformer (ViT)\cite{dosovitskiy2020image} following the standard ViT-base architecture with 12 Transformer layers, 12 attention heads, an embedding dimension of 768, and a hidden dimension of 3,072. The token size is 16 $\times$ 16, and learned absolute positional embeddings are added to each token. The backbone transforms images in the form of raw RGB pixel values to dense feature maps in a more semantically rich representation space learned from data. Each attentional pooler is responsible for computing a fixed number (denoted by $n$) of image tokens from the last layer representation of the ViT backbone using multi-headed attention and $n$ learned queries. For enabling cross-modal retrieval via contrastive learning, the first attentional pooler $f_\mathrm{contrast}(\,\cdot\,;\theta_{contrast})$ uses a single query ($n_\mathrm{contrast} = 1)$ to compute a single image token designed to capture the global representation of the image. The second attentional pooler $f_\mathrm{caption}(\,\cdot\,;\theta_\mathrm{caption})$ uses $n_\mathrm{caption} = 256$ queries to generate a set of 256 image tokens designed to capture more local and fine-grained details of the image, which are typically required for captioning. The text encoder and multimodal decoder are both GPT-style Transformer models that employ causal attention masks for left-to-right autoregressive language modeling. Similar to the image encoder, the text encoder and multimodal decoder consist of 12 Transformer layers with an embedding dimension of 768 and a hidden dimension of 3072. The text encoder includes an embedding table for mapping discrete word tokens to continuous embeddings and a set of learned absolute positional embeddings. Additionally, the text encoder appends a learned $\textless$\texttt{CLS}$\textgreater$ token to each tokenized caption, which has access to the full context during Transformer attention to extract a global representation of a given caption. The multimodal decoder inserts a cross-attention layer after each multiheaded self-attention layer for incorporating information from image tokens and includes a final language modeling head for predicting the distribution of the next token over the supported vocabulary. 

During visual-language pretraining, a mini-batch consists of $M$ image-caption pairs $(\textbf{x}_i, \textbf{w}_i)_{i=1}^{M}$, where $\textbf{w}_{i}$ = ($<$\texttt{BOS}$>$, $w_{i,1}, \ldots, w_{i,T}$, $<$\texttt{EOS}$>$) is a sequence of $T$ word tokens representing the $i$th caption. For a given pair ($\textbf{x}_i, \textbf{w}_i$), we let ($\textbf{u}_i, \textbf{v}_i$) be the output of $f_{\mathrm{contrast}}(\cdot;\theta_\mathrm{contrast})$ and the output of $g(\cdot;\theta)$ at the position corresponding to the $<$\texttt{CLS}$>$ token, respectively, after $\ell_2$-normalization. The complete objective is given by: 
\begin{equation}
\begin{split}
\mathcal{L} = &-\frac{1}{2M}\sum_{i=1}^{M} \log \frac{\exp \left(\tau \boldsymbol{u}_i^{T} \boldsymbol{v}_i\right)}{\sum_{j=1}^{M} \exp \left(\tau \boldsymbol{u}_i^{T}  \boldsymbol{v}_j\right)} -\frac{1}{2M}\sum_{j=1}^{M} \log \frac{\exp \left(\tau \boldsymbol{v}_j^{T}  \boldsymbol{u}_j\right)}{\sum_{i=1}^{M} \exp \left(\tau \boldsymbol{v}_j^{T}  \boldsymbol{u}_i\right)} \\ & -\frac{1}{M} \sum_{i=1}^{M}\sum_{t=1}^{T+1} \log p\left(w_{i,t} \mid w_{i, 0: t-1}, \mathbf{x}_i; \theta,\phi,\psi\right)
\end{split}
\end{equation}
The first and second terms represent image-to-text (\textit{i2t}) and text-to-image (\textit{t2i}) contrastive loss, respectively, to maximize the cosine-similarity scores between paired image and text embeddings relative to remaining negative pairings in the mini-batch. The last term seeks to maximize the log-likelihood of each observed token under the multimodal autoregressive language model (jointly parameterized by the image encoder, text encoder, and multimodal decoder), conditioned on previous tokens in the caption as well as the corresponding image. Each visual-language pretraining experiment was trained for 40 epochs, distributed across 8 NVIDIA A100 80GB GPUs with a local batch size of 48 per GPU, and uses gradient accumulation to achieve an effective global batch size of 1536. We set the image size to 448 $\times$ 448 px, where larger images are first resized along the shorter edge and center-cropped and smaller images are zero-padded as needed. For all optimization hyperparameters, refer to \textbf{Extended Data Table 29.}


\noindent\textbf{Pretraining unimodal encoders}\\
Prior work\cite{lu2023visual} has shown that before joint visual-language pretraining using paired image-caption data, performing self-supervised pretraining of unimodal modules using unpaired data can substantially improve downstream zero-shot transfer performance. We pre-train our image encoder using iBOT\cite{zhouimage}, a state-of-the-art, self-supervised pretraining algorithm for unlabeled image data. An in-house dataset of 16 million 256 $\times$ 256-sized image tiles are sampled and extracted at 20$\times$-equivalent magnification from the tissue regions of 21,442 WSIs spanning over 350 cancer subtypes under the OncoTree classification system\cite{kundra2021oncotree}. Detailed hyperparameters for image-only pretraining are provided in \textbf{Extended Data Table 30}. For pretraining the language model, we build a diverse corpus of pathology-relevant texts ranging from text from pathology educational texts and final diagnosis section of over 550k surgical pathology reports from Massachusetts General Hospital and over 400k select histopathology-relevant PubMed abstracts. We used regex to de-identify in-house diagnostic reports, notably replacing patient and physician names, specimen ids, medical record numbers, and dates with a corresponding special token in the vocabulary. We pretrain a 24-layer GPT-style autoregressive Transformer using the next-word prediction loss. Namely, given a sequence of word tokens $\textbf{w}$ = ($<$\texttt{BOS}$>$, $w_{i,1}, \ldots, w_{i,T}$, $<$\texttt{EOS}$>)$, we maximize the log-likelihood of each token under an autoregressive generative model parameterized by $\xi$:
\begin{equation}
   \mathcal{L}_\mathrm{clm}(\xi)= - \sum_{t=1}^{T+1} \log p\left(w_t \mid w_{0: t-1}; \xi \right) 
\end{equation}
Detailed hyperparameters for text-only pretraining are provided in \textbf{Extended Data Table 31}. After pretraining, the first 12 layers of the transformer-based language models and the embedding table are used to initialize the unimodal text encoder while the last 12 layers and the language modeling classifier head are used to initialize the corresponding parameters in the multimodal decoder. 

\noindent\textbf{Zero-shot transfer on ROIs/tiles}\\
For zero-shot transfer, we employ the method described in CLIP\cite{radford2021learning}. Each class is associated with a text prompt consisting of a class name (\textit{e.g.} ``\texttt{adenocarcinoma}") and a template (\textit{e.g.} ``\texttt{this is \{\}.}", see \textbf{Extended Data Table 34} for templates used across all tasks). For a prompt associated with class $j\in\{1,2,\dots,C\}$, we compute the $\ell_2$-normalized embedding $\mathbf{v}_j$ using a text encoder trained on our paired dataset to form the linear classifier weights. Since model performance can vary considerably depending on the choice of prompts, we measure the performance spread by sampling subsets from a pathologist-curated set of prompts and reporting the median. Alternatively, we can also ensemble all the prompts within a class by using the mean embedding over the prompts as the text embedding associated with that class. See \textbf{Extended Data Figure 2} for a comparison with and without ensembling. Analogously, for each image, we compute the $\ell_2$-normalized embedding $\mathbf{u}_i$. We then compute cosine-similarity scores between the image and each text embedding and the predicted class is consequently the class with the highest similarity score, \textit{i.e.},
\begin{equation}
    {\hat y}_i = \underset{j}{\text{argmax}}\;{\mathbf{u}_i}^T\mathbf{v}_j
\end{equation}
Since some evaluation sets are imbalanced, we report the balanced accuracy (\textit{i.e.,} the macro average over the accuracy obtained on each class) and the average $F1$ score weighted by the support of each class. For SICAP, we also report the quadratic Cohen's $\kappa$ score, which is often used for prostate Gleason grading\cite{silva2021self}, where errors between adjacent grading classes are penalized less.

Similarly for cross-modal retrieval, we use the same method as zero-shot classification above to retrieve the top-$K$ images that are closest in the aligned latent space to a specific text query (text-to-image retrieval). Image-to-text retrieval is performed analogously. To evaluate retrieval, we follow ALIGN\cite{jia2021scaling} and use Recall@$K$, \textit{i.e.}, for what percentage of the test set is the correct result in the top $K$ retrieved samples. We choose $K\in\{1,5,10\}$ and also report mean recall by averaging the scores over the three Recall@$K$'s.

Unless otherwise specified, we enforce the maximum image size to be 448 $\times$ 448 for CONCH via image resizing and center-cropping, similar to its pretraining configuration. For all models that are not our own, we use their provided processor function and default configuration for image and text processing in downstream evaluation.

\noindent\textbf{Extending zero-shot transfer to WSIs}\\
To extend zero-shot transfer to gigapixel images, we follow the method introduced by MI-Zero\cite{lu2023visual}. Namely, for classification over $C$ classes, the WSI is first divided into $N$ tiles and computes the $\ell_2$-normalized embeddings independently using the image encoder. For each tile embedding, we compute similarity scores with each text embedding following the method for tiles described above, obtaining a set $C$ similarity scores for each tile. To aggregate similarity scores across tiles, we use the top-$K$ pooling operator by averaging over the highest $K$ similarity scores for each class to obtain the slide-level similarity score. Consequently, the class with the highest slide-level score is the predicted class. We choose $K\in\{1,5,10,50,100\}$ and report metrics for the $K$ with the highest balanced accuracy for classification tasks and Cohen's $\kappa$ for DHMC LUAD. Similar to classification on tiles, we report slide-level balanced accuracy and weighted $F1$ score for classification tasks. For DHMC LUAD, since the task of LUAD subtyping can be subjective, we report Cohen's $\kappa$ score. 

We perform zero-shot slide-level segmentation using a similar approach as classification. We divide the WSI into tiles and compute similarity scores for each tile independently. However, instead of aggregating the scores across tiles into a single slide-level prediction, we map the tile-level scores to their corresponding spatial locations in the WSI, averaging the scores in overlapped regions. Finally for each pixel, we assign the class with the highest score as the prediction, producing 
a pixel-level segmentation mask. We compute the Dice score\cite{dice1945measures} to quantify the quality of the predicted segmentation mask relative to the ground truth. 

Details of WSI preprocessing for both classification and segmentation tasks are described in the \textbf{WSI processing} section.

\noindent\textbf{Supervised classification experiments}\\
We perform supervised classification experiments on all tasks with a labeled set of training examples available, including TCGA BRCA for BRCA subtyping, TCGA NSCLC for NSCLC subtyping, TCGA RCC for RCC subtyping, CRC100k for CRC tissue classification and SICAP for Gleason grading. For each dataset, we use the official train/test split if it is available or we use the remaining labeled cases for training after holding out the cases used for zero-shot classification evaluation (see \textbf{Downstream evaluation datasets} for a more detailed breakdown). For slide-level experiments, we consider 4 visual-language pretrained image encoders, including that of CONCH, PLIP, BiomedCLIP as well as OpenAICLIP. All 4 encoders follow the ViT-Base architecture with a patch size of 16 except PLIP, which uses a patch size of 32. For slide-level tasks, we additionally consider a ResNet50 encoder truncated after the 3rd residual block and with weights initialized from supervised classification on ImageNet, as it has been a common choice in weakly-supervised classification of WSIs. For ROI-level tasks, we add CTransPath\cite{wang2022transformer} as a baseline, which is a SOTA general purpose vision encoder trained with self-supervised learning on a large dataset of unlabeled histopathology images. The reason we do not use CTransPath for the slide-level tasks is that TCGA slides (including those used in our test sets), make up a large portion of the data used to train CTransPath and therefore may result in information leakage that unfairly inflate the performance of CTransPath on TCGA-based benchmarks. 

For all experiments, we standardize the image input size to 224 $\times$ 224. We use each image encoder to extract a low-dimensional feature embedding from each image (tiles in the case of WSIs). For CONCH, we use the output of the attentional pooler that corresponds to image / text alignment, with an embedding dimension of 512. For CLIP-based models including PLIP, BiomedCLIP and OpenAICLIP, we use the $\textless$\texttt{CLS}$\textgreater$ token, which is also used for image / text alignment during pretraining, and similarly has a dimension of 512. For ResNet50, we use global average pooling after the 3rd residual block to obtain an 1024-dimensional embedding. For CTransPath, we also use the $\textless$\texttt{CLS}$\textgreater$ token representation, which has an embedding dimension of 768. 

For WSI classification, we use the same preprocessing setup as zero-shot classification with MI-Zero. We use the widely used attention-based multiple instance learning (ABMIL)\cite{ilse2018attention} for weakly-supervised on WSIs using slide-level labels. The ABMIL model architecture consists of a fully-connected layer and ReLU non-linearity that first maps the inputs to an embedding dimension of 512, followed by a two-layer, gated variant (as described in the original paper) of the attention network, with a hidden dimension of 384. Lastly a fully-connected classifier head maps the attention-pooled slide-level representation to logits, which are interpreted as class probabilities after softmax normalization. We use dropout with $P = 0.25$ after each intermediate layer in the network for regularization. We train each model for 20 epochs on the training set, using an AdamW optimizer, a cosine learning rate scheduler and a learning rate of 1e-4. We use a weighted data sampler that increases the sampling probability of slides from minority classes such that on average the model sees the same number of slides from each class each epoch. The full set of hyperparameters is summarized in \textbf{Extended Data Table 32}. 

For ROI-level classification, we conduct linear probing by training a logistic regression model on top of the pretrained image embeddings of each encoder. We follow a practice  recommended by the large-scale self-supervised representation learning community\cite{kolesnikov2019revisiting} and set the $\ell_2$ regularization coefficient $\lambda$ to $\frac{100}{MC}$ where $M$ is the embedding dimension and $C$ is the number of classes. We use the lgbfs solver and set the maximum number of iterations to 800. 

For few-shot classification, we keep the test set the same, and vary the number of labeled examples per class for training (known as ``shot") from $n_c = 1,2,4,8,16,32,\ldots$ up to either $n_c = 512$ or the maximum number of labeled examples available for a given class. Otherwise, the hyperparamters and training setup remain the same as described above.

\noindent\textbf{Captioning with fine-tuning}\\
For captioning, we fine-tune the entire model on a small training set of image-caption pairs. When fine-tuning our setup, we set the weight for contrastive loss to zero. To evaluate performance, we report the commonly used metrics METEOR\cite{banerjee2005meteor} and ROUGE\cite{lin2004rouge}. For each model, we train for a maximum of 40 epochs and select the checkpoint with the highest METEOR on the validation set using an early-stopping patience of 10 epochs. At inference time, we generate captions using top-$K$ sampling\cite{fan2018hierarchical} as the decoding strategy with $K=50$, where at each time step, the $K$ most likely tokens are filtered and the probability mass is redistributed before sampling. Similar to zero-shot classification and retrieval, we set the maximum image size to 448 $\times$ 448. The full set of hyperparameters used to fine-tune captioning is presented in \textbf{Extended Data Table 33}.


\noindent\textbf{Evaluation metrics}\\
For classification tasks, we report balanced accuracy, weighted F1 score, and AUC ROC. \textbf{Balanced accuracy} is defined as the macro average of the recall of each class. \textbf{Weighted F1 score} is computed by taking the average of the F1 score (the harmonic mean of precision and recall) of each class, weighted by the support of each class. In the binary case, \textbf{AUC ROC} is the area under the receiver operating curve, which plots the true positive rate against the false positive rate as the classification threshold is varied. AUC ROC is generalized to the multiclass case by averaging over the AUC ROC of all pairwise combinations of classes. For retrieval, we use the metric \textbf{Recall@$K$}, which is the proportion of the data that is correctly retrieved among the top $K$ retrieved samples. Following ALIGN\cite{jia2021scaling}, we choose $K\in\{1,5,10\}$ and also compute the \textbf{mean recall}, which averages over the Recall@$K$'s. For segmentation, we report the \textbf{Dice score}, which is the same as the F1 score, and the precision and recall of the positive class. For captioning, we report METEOR and ROUGE for comparing the predicted caption with the ground truth caption. \textbf{METEOR}\cite{banerjee2005meteor} (Metric for Evaluation of Translation with Explicit ORdering) is a metric based on unigram matching that considers both precision and recall between the original and ground truth and takes into account synonyms and word forms. \textbf{ROUGE}\cite{lin2004rouge} (Recall-Oriented Understudy for Gisting Evaluation) is computes the overlap of $n$-grams between the predicted caption and ground truth. We use ROUGE-1, which considers unigrams. 

\noindent\textbf{Downstream evaluation datasets}\\
\textbf{Source A} is a dataset of image-caption pairs extracted from a held-out source from data scraping. We split multipanel figures and matched them with captions manually. Since we use this dataset for captioning as well, and the captions are generally noisy and often contain information not present in the images, a board-certified pathologist has cleaned the text and we use the cleaned version for all downstream tasks. After filtering and cleaning, we yield 797 images with an average width of 570 px and an average height of 428 px. We use this dataset in its entirety for cross-modal retrieval. We also use this dataset for captioning after performing a 70-10-20 split for training, validation, and testing. To avoid information leakage, the dataset split was performed at the figure level (taking into account multifigure panels that have been separated).\\
\textbf{Source B} is a dataset of image-caption pairs extracted from a held-out source from data scraping. Similar to Source A, we split multipanel figures and matched them with captions manually. After filtering and cleaning, we yield 1,755 images with an average width of 512 px and an average height of 410 px. Since the dataset is much bigger than Source A, we do not perform manual cleaning of the captions. We use this dataset for cross-modal retrieval.\\
\textbf{TCGA LUAD} consists of 165 image-caption pairs extracted from 49 LUAD H\&E histopathology slides from The Cancer Genome Atlas (TCGA)\footnote{\href{https://portal.gdc.cancer.gov}{portal.gdc.cancer.gov}}. 
For each slide, a board-certified pathologist chooses up to 5 tiles of interest from each slide and provides captions describing the tissue pattern as well as any notable morphological features. This yields a set of 165 image tiles with an average width of 656 px and average height of 642 px. We use this set of image tiles for cross-modal retrieval.\\
\textbf{TCGA BRCA} consists of invasive breast carcinoma (BRCA) H\&E FFPE diagnostic histopathology WSIs from TCGA. The dataset consist of cases for primary 
invasive ductal carcinoma (IDC) and invasive lobular carcinoma (ILC). After removing slides with missing metadata, we collected a total of 1,048 slides (837 IDC and 211 ILC). The \textbf{zero-shot test set} is a sampled subset of the full TCGA-RCC dataset consisting 150 WSIs (75 for each of class). For the supervised learning experiments, we hold out the zero-shot test set as the test set and use the rest of the slides as the \textbf{supervised training set} after excluding slides from patients who appear in the test set. This yields a training set of 881 slides (754 IDC, 127 ILC). See \textbf{Extended Data Table 35} for prompts used for each class in zero-shot classification.\\
\textbf{TCGA NSCLC} consists of non-small cell lung cancer (NSCLC) H\&E FFPE diagnostic histopathology WSIs from the TCGA. The dataset consists of cases for primary lung adenocarcinoma (LUAD) and lung squamous cell carcinoma (LUSC) cases. After removing slides with missing or incorrect metadata, we collected a total of 1,041 slides (529 LUAD and 512 LUSC). The \textbf{zero-shot test set} is a sampled subset of the full TCGA-RCC dataset consisting 150 WSIs (75 for each of class). For the supervised learning experiments, we hold out the zero-shot test set as the test set and use the rest of the slides as the \textbf{supervised training set} after excluding slides from patients who appear in the test set. This yields a training set of 846 slides (432 LUAD, 414 LUSC). See \textbf{Extended Data Table 35} for prompts used for each class in zero-shot classification.\\
\textbf{TCGA RCC} consists of renal cell carcinoma (RCC) H\&E FFPE diagnostic histopathology WSIs from the TCGA. The dataset consists of cases for primary clear cell renal cell carcinoma (CCRCC), papillary renal cell carcinoma (PRCC) and chromophobe renal cell carcinoma (CHRCC). After removing slides missing low-resolution downsamples, we collected a total of 922 WSIs (519 CCRCC, 294 PRCC and 109 CHRCC). The \textbf{zero-shot test set} is a sampled subset of the full TCGA-RCC dataset consisting 225 WSIs (75 for each of 3 classes). For the supervised learning experiments, we hold out the zero-shot test set as the test set and use the rest of the slides as the \textbf{supervised training set} after excluding slides from patients who appear in the test set. This yielded a training set of 693 slides (444 CCRCC, 215 PRCC, 34 ChRCC). See \textbf{Extended Data Table 35} for prompts used for each class in zero-shot classification.\\
\textbf{DHMC LUAD}\cite{wei2019pathologist} is a dataset of 143 H\&E LUAD slides, each labeled with the primary histologic growth pattern (59 Acinar, 51 Solid, 19 Lepidic, 9 Micropapillary, 5 Papillary). We only use this dataset for zero-shot classification. See \textbf{Extended Data Table 36} for prompts used for each class in zero-shot classification.\\
\textbf{CRC100k}\cite{kather2019predicting} is a dataset of 224 $\times$ 224 px image tiles at 0.5 microns per pixel (mpp) extracted from 50 patients with colorectal adenocarcinoma. Each image belongs to one of nine classes: adipose, background, debris, lymphocytes, mucus, smooth muscle, normal colon mucosa, cancer-associated stroma, and colorectal adenocarcinoma epithelium. For the \textbf{supervised dataset}, we use the officially provided splits of 100,000 images in the train set and 7,180 images in the test set. For the \textbf{zero-shot test set}, we only use the official test set. We refer the reader to \textbf{Extended Data Table 37} for prompts used for each class in zero-shot classification.\\
\textbf{WSSS4LUAD}\cite{han2022wsss4luad} is a dataset of lung adenocarcinoma image tiles of around 200 to 500 px in dimension each labeled as tumor, tumor-associated stroma, and/or normal. For our evaluation, we filter for the samples that have only one ground truth label. We are left with 4,693 images from the official training split. Refer to \textbf{Extended Data Table 38} for prompts used for each class in zero-shot classification.\\
\textbf{SICAP}\cite{silva2021self} consists of 512 $\times$ 512 px images extracted from 155 WSIs of core-needle biopsies of prostate cancer, digitized at 10$\times$ magnification. The official training and testing split partitions the dataset into 9,959 images from 124 WSIs for training, and 2,122 images from 31 WSIs for testing. Each tile is labeled by the primary Gleason pattern (3, 4, or 5) or as non-cancerous (NC). For zero-shot classification, we use the official test set for evaluation only while for supervised classification, we use the official splits for training and testing. For zero-shot segmentation (tumor \textit{vs.} benign), we use the slides from the official test split and corresponding pixel-level segmentation mask for evaluation (combining Gleason patterns 3, 4, and 5 as the tumor class). Refer to \textbf{Extended Data Table 38} for prompts used for each class in zero-shot classification and segmentation.\\
\textbf{DigestPath}\cite{da2022digestpath} is a dataset of 660 colonoscopy H\&E tissue section images from 324 patients, acquired at 20$\times$-equivalent magnification. We use the subset of 250 images from 93 patients for which pixel-level lesion annotation for colorectal cancer tissue is provided and perform zero-shot segmentation evaluation. Refer to \textbf{Extended Data Table 38} for prompts used for each class in zero-shot segmentation.

\noindent\textbf{WSI processing}\\
For slide-level tasks, the processing pipeline for WSIs consist of tissue segmentation, tiling, and feature extraction. We use the CLAM library\cite{lu2021data} for tissue segmentation, which computes a binary mask for tissue using binary thresholding along the saturation channel after converting a downsample of the slide form the RGB to HSV color space. Median blurring and morphololgical closing were used to smooth tissue contours and remove artifacts. The contours are filtered by area to yield the segmentation mask. For zero-shot and supervised classification, we follow previous conventions\cite{wang2022transformer, lu2021data} and divide the segmented tissue regions into contiguous 256 $\times$ 256 px tiles at 10$\times$-equivalent magnification. For segmentation, we extract tiles using a smaller tile size (224 $\times$ 224 px) with 75\% overlap at the highest magnification possible (\textit{i.e.}, 10$\times$ for SICAP and 20$\times$ for DigestPath) in order to achieve more fine-grained predictions. After tiling, for feature extraction, we resize all tiles to 224 $\times$ 224 px and compute embeddings for each tile independently using a frozen pretrained image encoder and cache them for downstream evaluation. 

\noindent\textbf{Pretraining dataset characterization}\\
We estimate the distribution of topics covered by our pretraining captions. We first create a list of 19 topics that cover major anatomic sites relevant to the study of pathology. For each topic, a board-certified pathologist then curates a list of keywords associated with the topic. We then map a caption to a topic if it contains a specific word. Since it is impractical to curate an exhaustive set of keywords to cover all captions, we use $k$-nearest neighbors (kNN) with $k=5$ to categorize the remaining captions. The distribution of captions on the topics are shown in \textbf{Figure 1b}. Within each topic (as well as the overall dataset), we qualitatively visualize the contents of the captions using wordclouds (\textbf{Extended Data Figure 1}). 

\noindent\textbf{Statistical analysis}\\
Nonparametric bootstrapping with 1,000 samples is used to construct 95\% confidence intervals for model performance. Observed differences in model performance were tested for statistical significance via two-sided paired permutation test with 1,000 permutations. In each permutation, independent predictions or prediction outcomes of two models are randomly swapped to obtain a new difference in model performance. The p-value is the proportion of differences in model performance that are greater than the observed difference in terms of absolute value. For comparing zero-shot classification performance in the setting of randomly sampled prompts (\textit{i.e.}, no prompt ensembling), the null hypothesis is that there is no difference in the median performance of two models over the same set of sampled prompts. For all other tests, the null hypothesis is that there is no difference in the model performance for the given test set. 




\noindent\textbf{Computing hardware and software}\\
We used Python (version 3.8.13) for all experiments and analyses in the study, which can be replicated using open-source libraries as outlined below. For task-agnostic pretraining, we used 8$\times$80GB NVIDIA A100 GPUs configured for multi-GPU training using distributed data-parallel (DDP) as implemented by the popular open source deep learning framework PyTorch (version 2.0.0, CUDA 11.7) (\href{https://pytorch.org/}{pytorch.org}). All downstream experiments were conducted on single 24GB NVIDIA 3090 GPUs. For unimodal pretraining of our visual encoder using iBOT, we modify the vision transformer implementation maintained by the open-source Timm library (version 0.9.2) from Hugging Face (\href{https://huggingface.co/}{huggingface.co}) for the encoder backbone and use the original iBOT implementation (\href{https://github.com/bytedance/ibot}{github.com/bytedance/ibot}) for training. For NLP workflows, we used open-source libraries provided by Hugging Face. Notably, we used Transformers (version 4.27.3) and Accelerate (version 0.15.0) for tokenization of text data and unimodal pretraining of our language model, and Evaluate (version 0.4.0) for its implementation of common machine translation/image captioning metrics including ROUGE and METEOR. We integrate our pretrained unimodal visual encoder and language model into the open clip library (version 2.14.0) for visual-language pretraining using the CoCa framework. All WSI processing was supported by OpenSlide (version 4.3.1) and openslide-python (version 1.2.0). We use Scikit-learn (version 1.2.1) for its implementation of common machine learning model evaluation metrics for image classification and to train logistic regression models for linear probe experiments. Implementations of other visual-language models benchmarked in the study are found on Hugging Face model hub (\href{https://huggingface.co/models}{huggingface.co/models}): PLIP (\href{https://huggingface.co/vinid/plip}{vinid/plip}), BiomedCLIP (\href{https://huggingface.co/microsoft/BiomedCLIP-PubMedBERT_256-vit_base_patch16_224}{microsoft/BiomedCLIP-PubMedBERT\_256-vit\_base\_patch16\_224}, OpenAICLIP (\href{https://huggingface.co/openai/clip-vit-base-patch16}{openai/clip-vit-base-patch16}), GIT-base (\href{https://huggingface.co/microsoft/git-base}{microsoft/git-base}), GIT-large (\href{https://huggingface.co/microsoft/git-large}{microsoft/git-large}). Pillow (version 9.3.0) and Opencv-python were used to perform basic image processing tasks . Matplotlib (version 3.7.1) and Seaborn (version 0.12.2) were used to create plots and figures. Usage of other miscellaneous Python libraries is listed in the \textbf{Reporting Summary}.

\noindent\textbf{\large{Data availability}}\\
TCGA whole slide data and labels are available from the NIH genomic data commons (\href{https://portal.gdc.cancer.gov}{portal.gdc.cancer.gov}). DHMC LUAD whole slide data and labels can be accessed through the Dartmouth Biomedical Informatics Research and Data Science website (\href{https://bmirds.github.io/LungCancer/}{bmirds.github.io/LungCancer/}). SICAP whole slide and tile data with corresponding labels can be accessed through the data portal at \href{https://data.mendeley.com/datasets/9xxm58dvs3/1}{data.mendeley.com/datasets/9xxm58dvs3/1}. CRC100k tile data and labels can be found at \href{https://zenodo.org/record/1214456}{zenodo.org/record/1214456}. WSSS4LUAD image tiles and labels can be found at \href{https://wsss4luad.grand-challenge.org/}{wsss4luad.grand-challenge.org/}. 
Pretraining data was curated from image-caption pairs in educational resources and PubMed. Educational resources are subject to copyright terms of publishers and will not be made available. The unprocessed PubMed Central Open Access dataset are available from the NIH PubMed Central website (\href{https://www.ncbi.nlm.nih.gov/pmc/tools/openftlist/}{ncbi.nlm.nih.gov/pmc/tools/openftlist/}). Pathology reports used in unimodal text pretraining are in-house data used with institutional permission through IRB approval for the current study and are thus not publicly available. All requests for data collected or curated in-house will be evaluated based on institutional and departmental policies to determine whether the data requested is subject to intellectual property or patient privacy obligations. 

\noindent\textbf{\large{Code availability}} \\
Code for performing various downstream tasks using a pretrained visual-language foundation model will be released upon publication. Model weights depend on the use of proprietary patient data and may be requested upon institutional permission and case by case approval. We have documented all technical deep learning methods and software libraries used in the study while ensuring the paper is accessible to the broader clinical and scientific audience. 

\noindent\textbf{\large{Author contributions}}\\
F.M., M.Y.L., B.C., and D.F.K.W. conceived the study and designed the experiments. M.Y.L., B.C., R.J.C., T.D., I.L., D.F.K.W, I.O. and L.P.L. performed data collection and cleaning for data used for unimodal and visual-language pretraining. M.Y.L, B.C. and R.J.C performed model development. M.Y.L., B.C., D.F.K.W. and G.J. performed experimental analysis. M.Y.L., B.C., D.F.K.W, A.Z., R.J.C., I.L., T.D., G.J., F.M., G.G, L.P.L and A.V.P. interpreted experimental results and provided feedback on the study. M.Y.L., B.C., D.F.K.W. and F.M. prepared the manuscript with input from all co-authors. F.M. supervised the research.

\noindent\textbf{\large{Acknowledgements}}\\
We thank Andrew Song for his feedback. This work
was supported in part by the BWH president’s fund, BWH
\& MGH Pathology, and NIGMS R35GM138216 (F.M.).
M.Y.L. was also supported by the Siebel Scholars program.
R.J.C. was also supported by the NSF Graduate Fellowship.
\end{spacing}

\newpage

\begin{figure*}
\includegraphics[width=\textwidth]{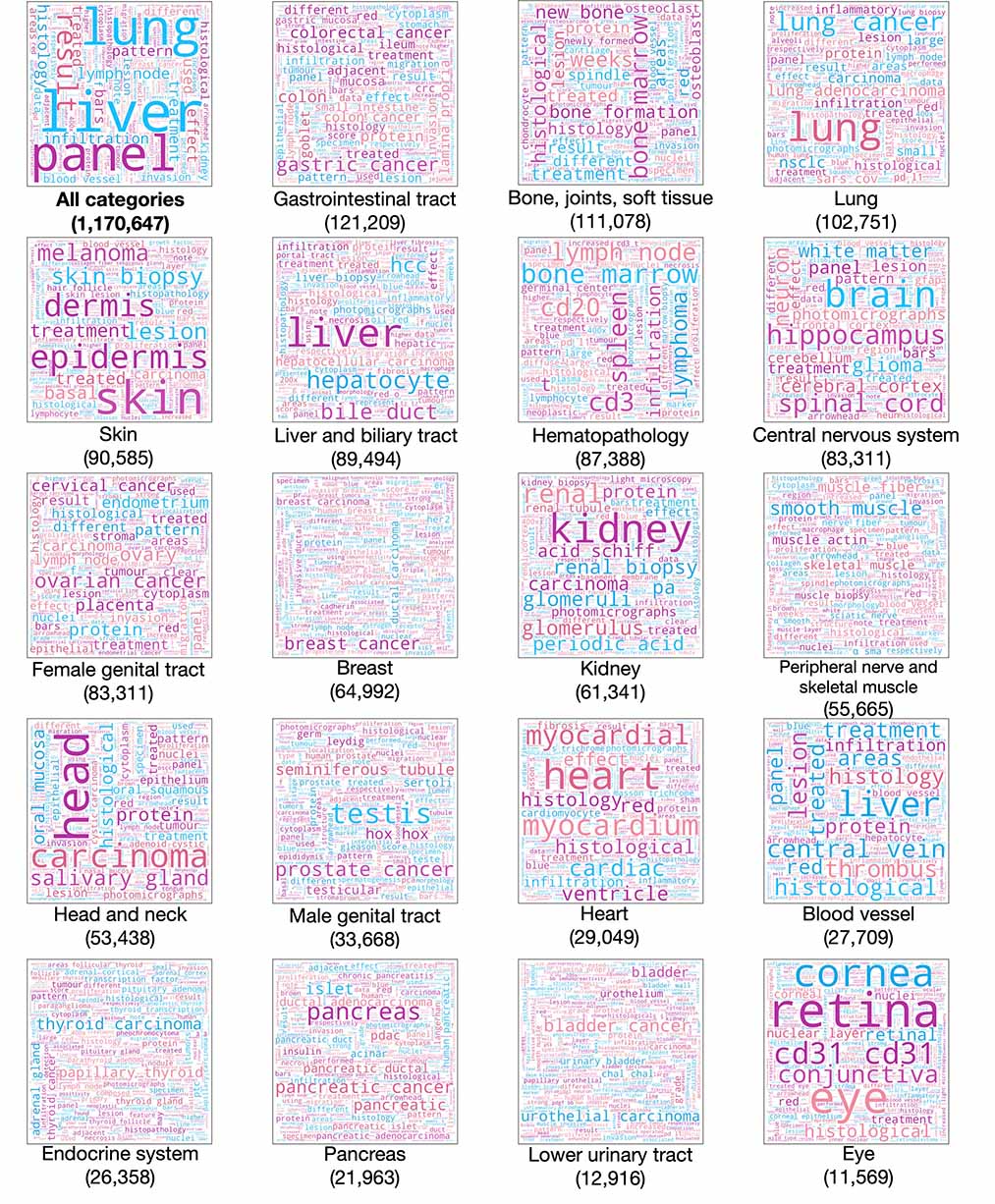}
\caption*{\textbf{Extended Figure 1: Caption content of pre-training dataset.} Wordclouds of captions to qualitatively visualize the caption content of each category in the pre-training dataset. Larger words are more represented in the captions. Common articles, nouns, and verbs are ignored.}
\end{figure*}

\clearpage
\begin{figure*}
\centering
\includegraphics[width=0.95\textwidth]{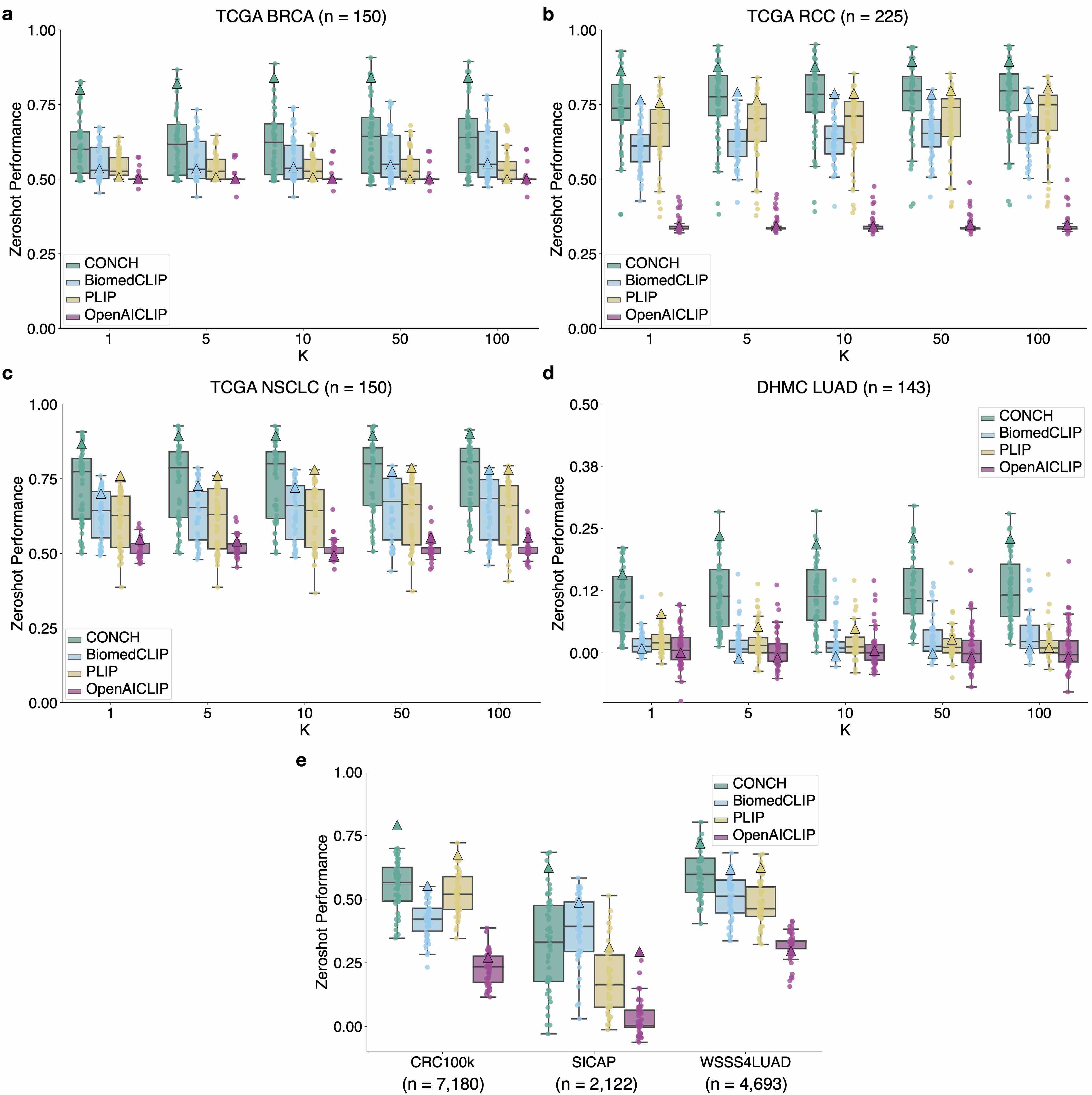}
\caption*{\textbf{Extended Figure 2: Zero-shot classification: single prompt vs. ensembling.} \textbf{a-d}, slide-level tasks. \textbf{e}, ROI-level tasks. We compare using a single text prompt per class \textit{vs.} ensembling over multiple class names and templates. Since zero-shot performance of a visual-language pretrained model can be sensitive to the prompts used\cite{radford2021learning}, when using a single prompt per class, for each class, we independently randomly sample a prompt from the pool of candidate templates and class names (see \textbf{Extended Data Tables 34-38} for the prompt pools). We randomly sample 50 sets of prompts for each task, and plot the resulting distribution of zero-shot performance for each model using boxplot. Each dot corresponds to a single set of prompts ($n=50$ for each box). Boxes indicate quartile values and whiskers extend to data points within 1.5$\times$ the interquartile range. Triangles indicate the performance of prompt ensembling. For slide-level tasks we show performance for all $K$s used in top-$K$ pooling. We observe prompt ensembling can substantially boost performance (relative to the median performance of randomly sampled single prompts) for most models in most tasks, except when the median performance is near random chance, such as for OpenAICLIP on most tasks and PLIP on TCGA BRCA. The poor median performance in these scenarios indicate that the model fails to perform under the majority of prompts sampled and therefore it is unsurprising that the ensembled prompt perform equally bad or worse. See \textbf{Extended Data Tables 1-14} for more results.}
\end{figure*}

\clearpage
\begin{figure*}
\centering
\includegraphics[width=.85\textwidth]{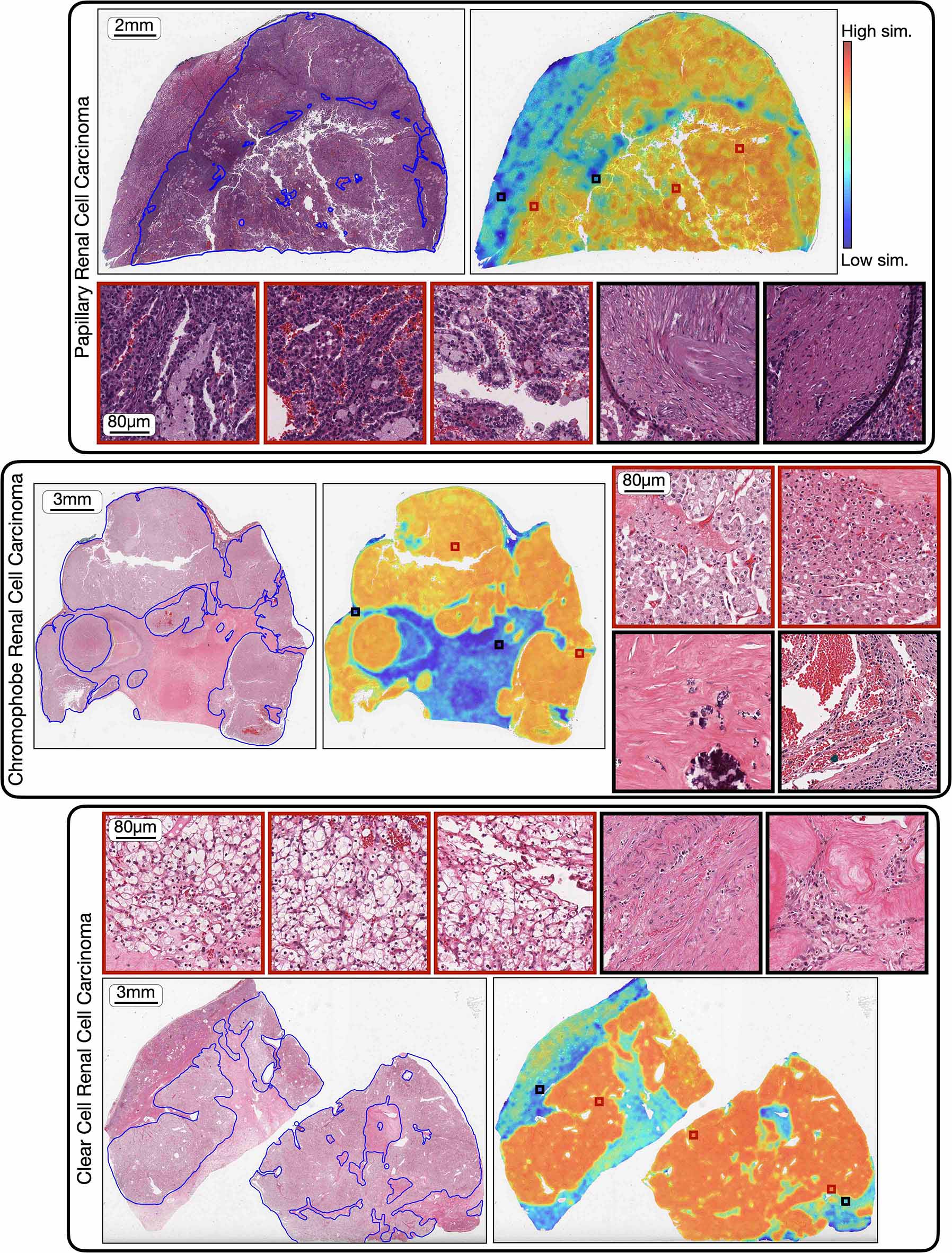}
\caption*{\textbf{Extended Figure 3: CONCH heatmaps, renal cell carcinomas.} Pathologist-annotated H\&E images, corresponding cosine-similarity heatmaps of, from top to bottom, papillary, chromophobe, and clear cell renal cell carcinomas. Tiles of high similarity (red border) and low similarity (black border) with the predicted class label are randomly sampled and displayed next to each heatmap. We find excellent agreement between the annotated image and the regions of the slide with high similarity, with the tiles demonstrating stereotypical morphology of the tumors within the high-similarity regions and stroma or other normal constituents of the kidney in the low similarity regions.}
\end{figure*}

\clearpage
\begin{figure*}
\centering
\includegraphics[width=.9\textwidth]{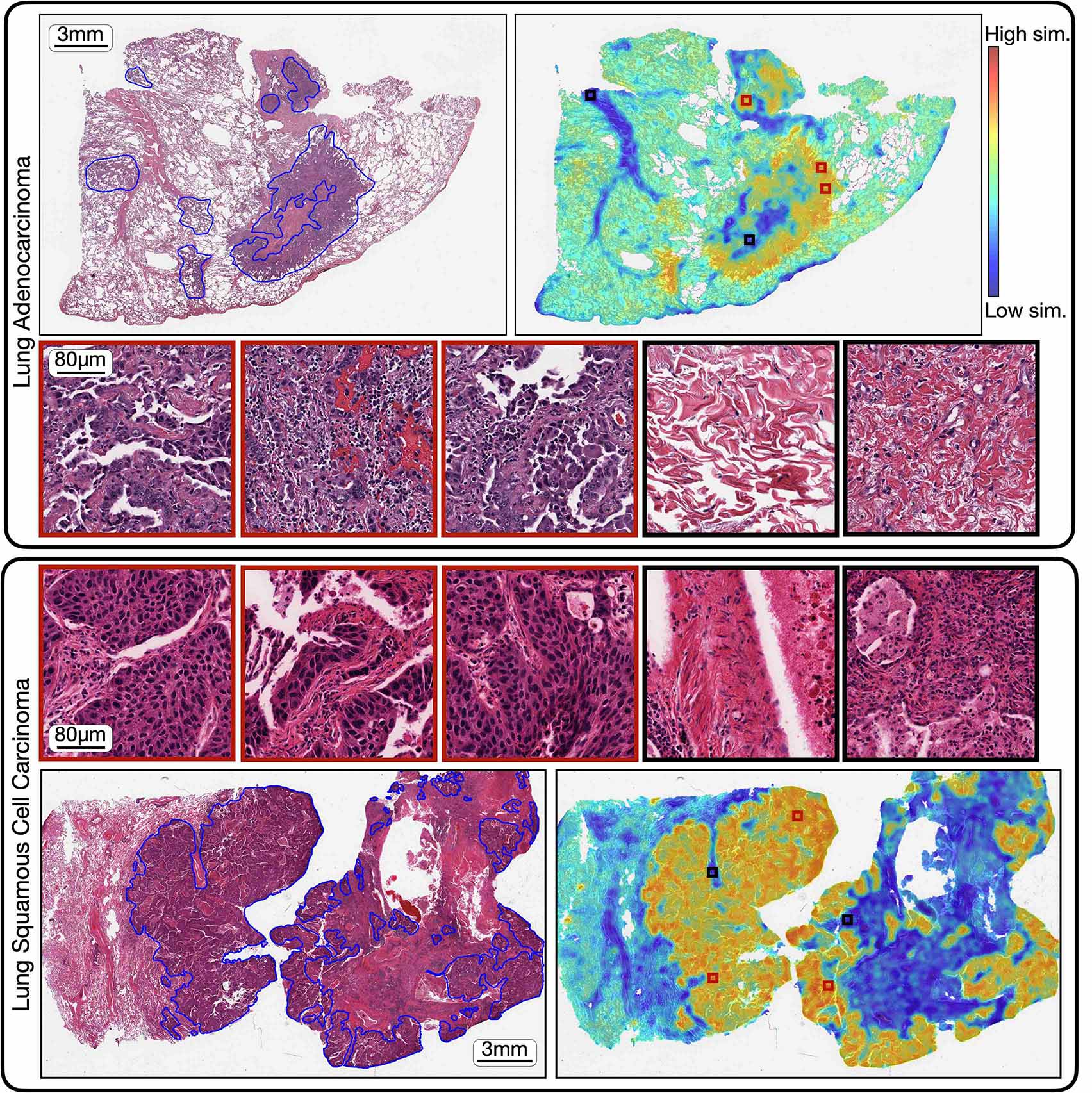}
\caption*{\textbf{Extended Figure 4: CONCH heatmaps, non-small cell lung carcinomas.} Pathologist-annotated H\&E images, corresponding cosine-similarity heatmaps of adenocarcinoma (top) and squamous cell carcinoma (bottom) of the lung. Tiles of high similarity (red border) and low similarity (black border) with the predicted class label are randomly sampled and displayed next to each heatmap. We find excellent agreement between the annotated image and the regions of the slide with high similarity, with the tiles demonstrating stereotypical morphology of the tumors within the high-similarity regions and stroma or other normal constituents of the lung in the low similarity regions.}
\end{figure*}

\clearpage
\begin{figure*}
\centering
\includegraphics[width=.9\textwidth]{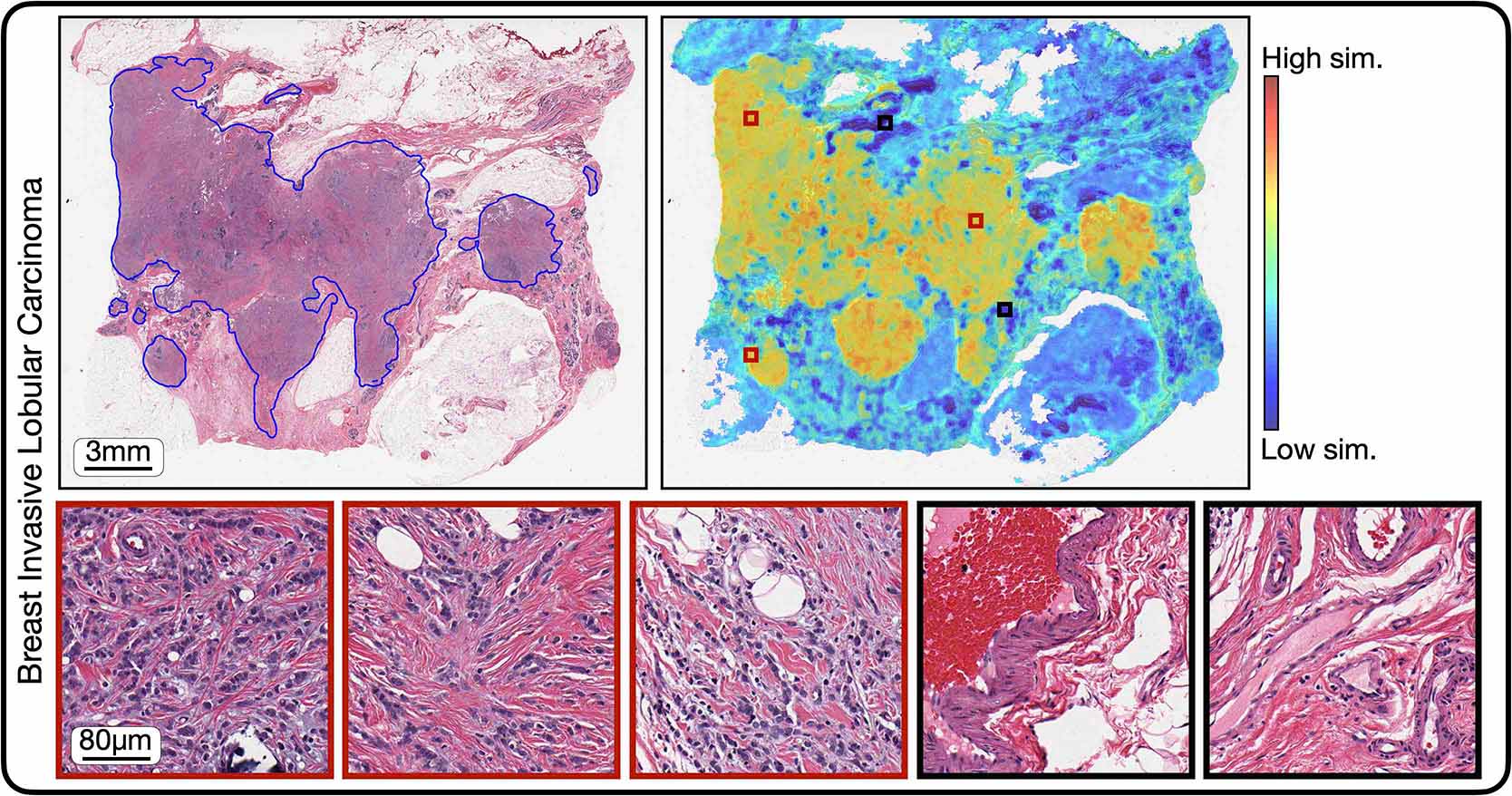}
\caption*{\textbf{Extended Figure 5: CONCH heatmap, lobular carcinoma of the breast.} Pathologist-annotated H\&E image, corresponding cosine-similarity heatmap of lobular carcinoma of the breast. Tiles of high similarity (red border) and low similarity (black border) with the predicted class label are randomly sampled and displayed next to the heatmap. As with the ductal carcinoma heatmap in \textbf{Fig. 2e}, we find excellent agreement between the annotated image and the regions of the slide with high similarity, with the tiles demonstrating stereotypical morphology of lobular caricnoma within the high-similarity regions and stroma or other normal constituents of the breast in the low similarity regions.}
\end{figure*}

\clearpage

\clearpage
\begin{figure*}
\centering
\includegraphics[width=\textwidth]{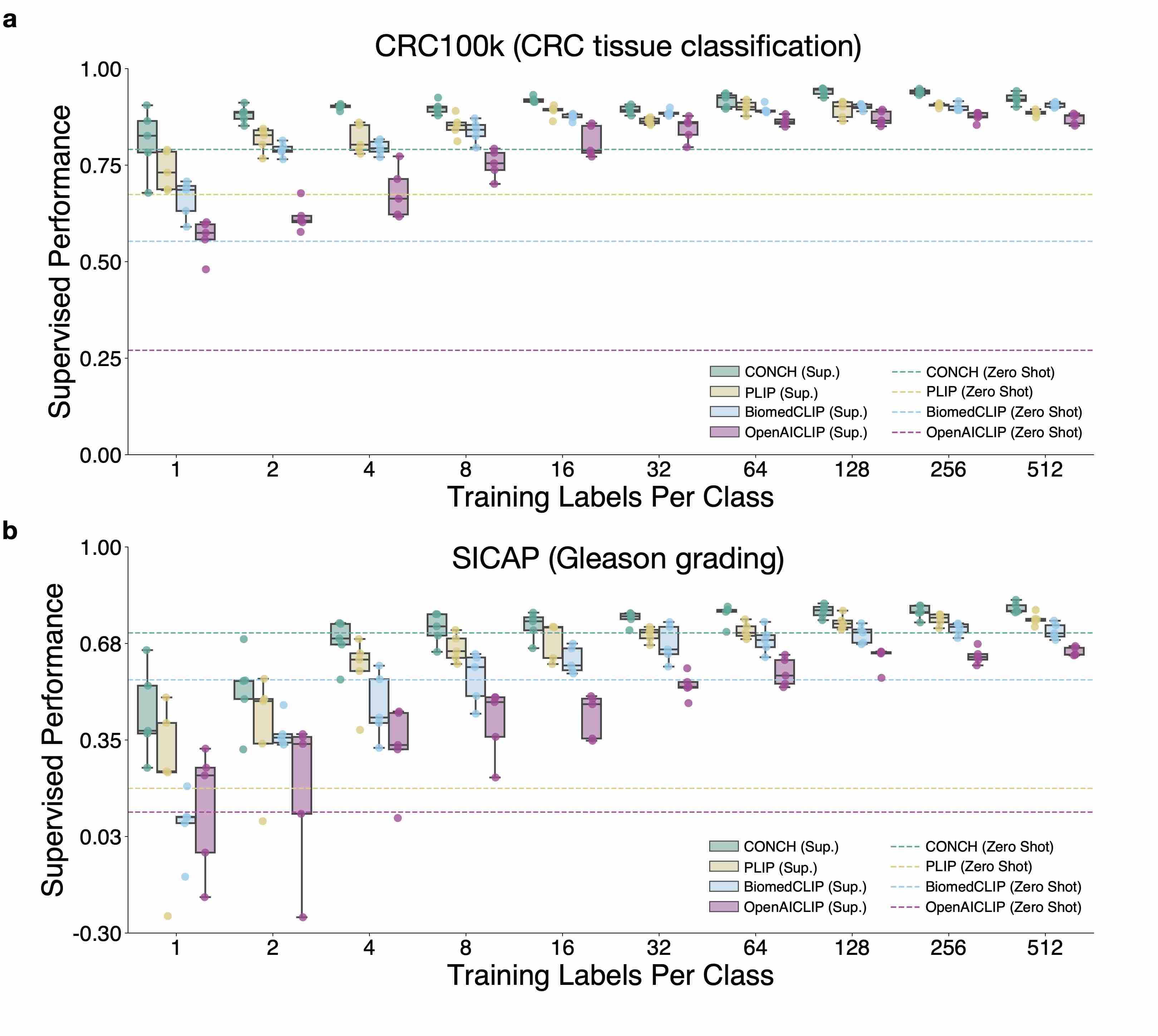}
\caption*{\textbf{Extended Figure 6: ROI-level few-shot classification experiments.} We investigate the label efficiency of different visual-language pretrained encoders in the few-shot setting where we vary the number of training labels per class ($n_{c}$), for $n_{c} = 1, 2, 4, 8, 16 \ldots$ up to 512. For each $n_c$, we sample 5 different sets of training examples and perform linear probing on each training set using associated image labels (see \textbf{Supervised classification experiments} for details). We show their individual model performance via boxplot (\textit{i.e.}, $n=5$ for each box) to study the variance in model performance when performing supervised learning with very few training examples. Boxes indicate quartile values and whiskers extend to data points within 1.5$\times$ the interquartile range. For reference, the zero-shot performance of each model is shown as a dotted line on the same plot. In terms of few-shot supervised learning, CONCH achieves better performance (\textit{i.e.} in terms of the median accuracy of 5 runs) than other encoders for different sizes of training set and for all tasks. Additionally,in SICAP, we find CONCH zero-shot performance to be competitive with PLIP and BiomedCLIP few-shot up to 64 labels per class. 
}
\end{figure*}

\clearpage
\begin{figure*}
\includegraphics[width=\textwidth]{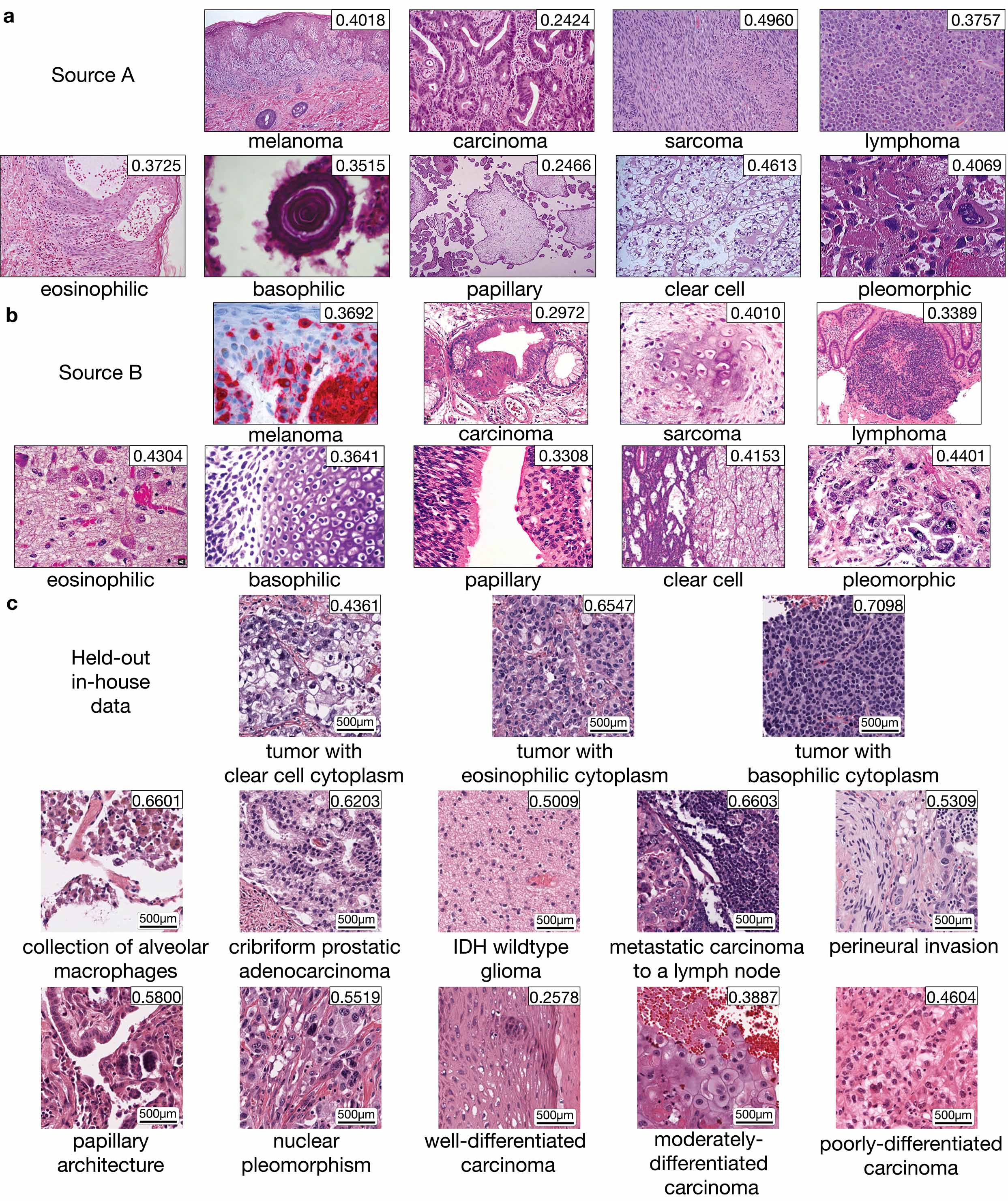}
\caption*{\textbf{Extended Figure 7: Additional Retrieval Examples.} \textbf{a.} Example images among top-10 retrieved results for each prompt from Source A. Similarity scores between each image and prompt are shown in the top-right corner of each image. Magnification information is unknown. \textbf{b.} Examples among top-10 results for each prompt from Source B. Magnification information is unknown. \textbf{c.} Retrieved examples (among top-10) using complex prompts with detailed morphological information. Images are from an in-house dataset of tiles sampled from 1,620 cases held-out during pretraining, spanning 108 OncoTree codes (5 for each code).}
\end{figure*}

\clearpage
\begin{figure*}
\includegraphics[width=\textwidth]{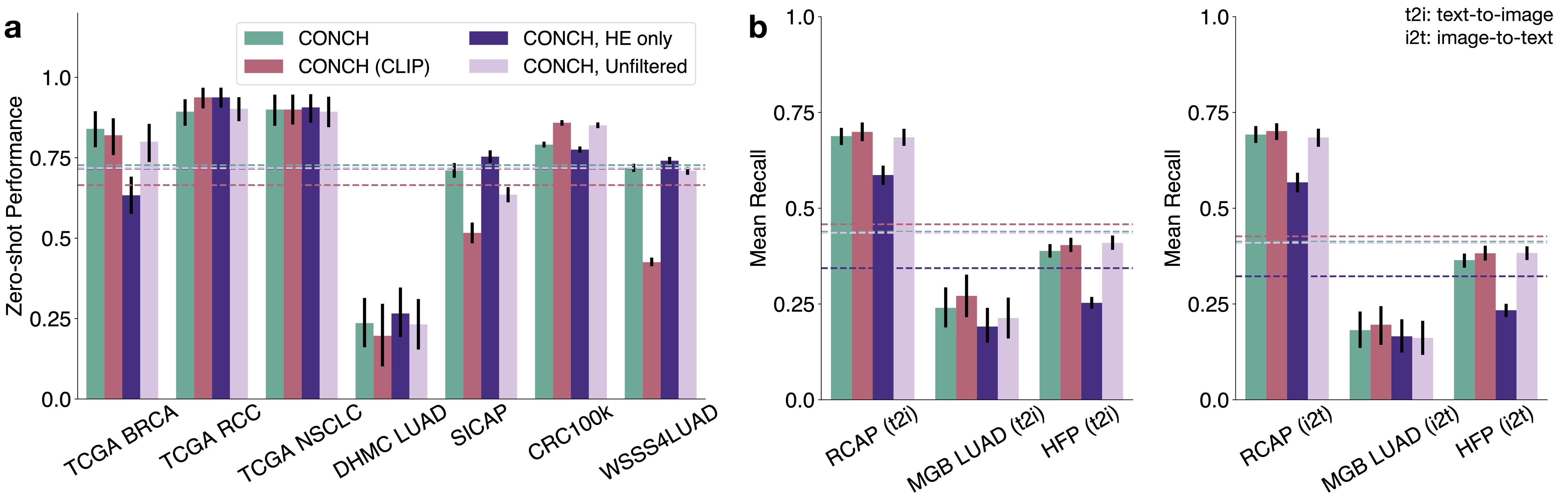}
\caption*{\textbf{Extended Figure 8: CONCH Pretraining Data Ablations.} Comparison between CONCH pretrained on human-only data ($n=1,170,647$) using CoCa \textit{vs.} human-only data using CLIP \textit{vs.} H\&E only data ($n=457,372$) \textit{vs.} the full unfiltered dataset ($n=1,786,362$). \textbf{a.} Zero-shot performance on downstream subtyping (TCGA BRCA, $n = 150$; TCGA RCC, $n = 225$; TCGA NSCLC, $n = 150$; DHMC LUAD, $n = 143$; CRC100k, $n = 7,180$; WSSS4LUAD, $n = 4,693$) and grading (SICAP, $n = 2,122$) tasks. Following pre-established conventions, Cohen’s $\kappa$ is reported for SICAP and quadratically weighted Cohen’s $\kappa$ is reported for DHMC LUAD, while balanced accuracy is reported for all other tasks. CONCH performs the best on average. \textbf{b.} Model performance in cross-modal retrieval on 3 datasets of image-text pairs (Source A, $n=797$; Source B, $n=1,755$; TCGA LUAD, $n=165$). CONCH (CLIP) performs the best on average.}
\end{figure*}

\clearpage
\begin{table}
\centering
\begin{tabular}{llll}
\toprule
       Model name &    Balanced accuracy &          Weighted F1 &              ROC AUC \\
\midrule
            CONCH & \textbf{0.840 (0.783, 0.895)} & \textbf{0.839 (0.775, 0.893)} & \textbf{0.932 (0.886, 0.970)} \\
       PLIP & 0.507 (0.500, 0.523) & 0.348 (0.263, 0.438) & 0.688 (0.600, 0.771) \\
       BiomedCLIP & 0.553 (0.510, 0.600) & 0.465 (0.358, 0.557) & 0.852 (0.784, 0.912) \\
       OpenAICLIP & 0.500 (0.500, 0.500) & 0.333 (0.255, 0.418) & 0.643 (0.544, 0.726) \\
\bottomrule
\end{tabular}
\caption{\textbf{Zero-shot classification with prompt ensembling on BRCA subtyping} ($n=150$) in terms of balanced accuracy, weighted F1 score, and ROC AUC. Best performing model for each metric is bolded. 95\% CI is included in parentheses.}
\end{table}

\begin{table}
\centering
\begin{tabular}{llll}
\toprule
       Model name &    Balanced accuracy &          Weighted F1 &              ROC AUC \\
\midrule
            CONCH & \textbf{0.893 (0.850, 0.932)} & \textbf{0.895 (0.852, 0.934)} & \textbf{0.973 (0.952, 0.991)} \\
             PLIP & 0.804 (0.755, 0.850) & 0.797 (0.740, 0.850) & 0.956 (0.932, 0.975) \\
       BiomedCLIP & 0.791 (0.737, 0.840) & 0.789 (0.733, 0.842) & 0.924 (0.893, 0.950) \\
       OpenAICLIP & 0.347 (0.333, 0.363) & 0.194 (0.141, 0.252) & 0.673 (0.626, 0.719) \\
\bottomrule
\end{tabular}
\caption{\textbf{Zero-shot classification with prompt ensembling on RCC subtyping} ($n=225$) in terms of balanced accuracy, weighted F1 score, and ROC AUC. Best performing model for each metric is bolded. 95\% CI is included in parentheses.}
\end{table}

\begin{table}
\centering
\begin{tabular}{llll}
\toprule
       Model name &    Balanced accuracy &          Weighted F1 &              ROC AUC \\
\midrule
            CONCH & \textbf{0.900 (0.849, 0.946)} & \textbf{0.900 (0.852, 0.947)} & \textbf{0.964 (0.936, 0.985)} \\
             PLIP & 0.787 (0.720, 0.849) & 0.786 (0.720, 0.847) & 0.838 (0.768, 0.899) \\
       BiomedCLIP & 0.780 (0.713, 0.843) & 0.776 (0.704, 0.840) & 0.877 (0.819, 0.922) \\
       OpenAICLIP & 0.553 (0.503, 0.604) & 0.478 (0.382, 0.570) & 0.603 (0.514, 0.689) \\
\bottomrule
\end{tabular}
\caption{\textbf{Zero-shot classification with prompt ensembling on NSCLC subtyping} ($n=150$) in terms of balanced accuracy, weighted F1 score, and ROC AUC. Best performing model for each metric is bolded. 95\% CI is included in parentheses.}
\end{table}

\begin{table}
\centering
\begin{tabular}{llll}
\toprule
       Model name &        Cohen's $\kappa$ &    Balanced accuracy &          Weighted F1 \\
\midrule
            CONCH & \textbf{0.236 (0.161, 0.314)} & \textbf{0.453 (0.332, 0.556)} & \textbf{0.324 (0.242, 0.413)} \\
             PLIP & 0.079 (0.014, 0.142) & 0.326 (0.250, 0.404) & 0.198 (0.129, 0.267) \\
       BiomedCLIP & 0.009 (0.000, 0.041) & 0.259 (0.204, 0.327) & 0.032 (0.005, 0.071) \\
       OpenAICLIP & 0.004 (0.000, 0.021) & 0.204 (0.200, 0.214) & 0.045 (0.015, 0.087) \\
\bottomrule
\end{tabular}
\caption{\textbf{Zero-shot classification with prompt ensembling on DHMC LUAD} ($n=143$) in terms of Cohen's $\kappa$, balanced accuracy, and weighted F1 score. Best performing model for each metric is bolded. 95\% CI is included in parentheses.}
\end{table}

\begin{table}
\centering
\begin{tabular}{llll}
\toprule
       Model name & Quadratic weighted $\kappa$ &    Balanced accuracy &          Weighted F1 \\
\midrule
            CONCH &     \textbf{0.711 (0.688, 0.734)} & \textbf{0.624 (0.609, 0.640)} & 0.424 (0.401, 0.449) \\
             PLIP &     0.187 (0.153, 0.220) & 0.312 (0.297, 0.326) & 0.099 (0.087, 0.113) \\
       BiomedCLIP &     0.553 (0.520, 0.585) & 0.487 (0.468, 0.505) & \textbf{0.437 (0.413, 0.459)} \\
       OpenAICLIP &     0.107 (0.081, 0.134) & 0.294 (0.280, 0.309) & 0.193 (0.177, 0.211) \\
\bottomrule
\end{tabular}
\caption{\textbf{Zero-shot classification with prompt ensembling on SICAP} ($n=2,122$) in terms of quadratic weighted Cohen's $\kappa$, balanced accuracy, and weighted F1 score. Best performing model for each metric is bolded. 95\% CI is included in parentheses.}
\end{table}

\begin{table}[]
\centering
\begin{tabular}{llll}
\toprule
       Model name &    Balanced accuracy &          Weighted F1 &              ROC AUC \\
\midrule
            CONCH & \textbf{0.791 (0.782, 0.800)} & \textbf{0.803 (0.794, 0.813)} & \textbf{0.979 (0.977, 0.981)} \\
             PLIP & 0.674 (0.665, 0.683) & 0.687 (0.676, 0.698) & 0.944 (0.941, 0.947) \\
       BiomedCLIP & 0.553 (0.542, 0.564) & 0.533 (0.521, 0.545) & 0.924 (0.921, 0.928) \\
       OpenAICLIP & 0.271 (0.262, 0.280) & 0.247 (0.236, 0.258) & 0.781 (0.777, 0.786) \\
\bottomrule
\end{tabular}
\caption{\textbf{Zero-shot classification with prompt ensembling on CRC100k} ($n=7,180$) in terms of balanced accuracy, weighted F1 score, and ROC AUC. Best performing model for each metric is bolded. 95\% CI is included in parentheses.}
\end{table}

\begin{table}
\centering
\begin{tabular}{llll}
\toprule
       Model name &    Balanced accuracy &          Weighted F1 &              ROC AUC \\
\midrule
            CONCH & \textbf{0.719 (0.706, 0.731)} & \textbf{0.705 (0.692, 0.718)} & \textbf{0.877 (0.870, 0.885)} \\
             PLIP & 0.624 (0.609, 0.638) & 0.636 (0.622, 0.650) & 0.790 (0.780, 0.801) \\
       BiomedCLIP & 0.616 (0.603, 0.628) & 0.575 (0.559, 0.589) & 0.824 (0.815, 0.833) \\
       OpenAICLIP & 0.296 (0.290, 0.303) & 0.196 (0.183, 0.209) & 0.496 (0.484, 0.508) \\
\bottomrule
\end{tabular}
\caption{\textbf{Zero-shot classification with prompt ensembling on WSSS4LUAD} ($n=4,693$) in terms of balanced accuracy, weighted F1 score, and ROC AUC. Best performing model for each metric is bolded. 95\% CI is included in parentheses.}
\label{tab:my_label}
\end{table}

\newpage
\begin{table}
\centering
\begin{tabular}{llll}
\toprule
       Model name &    Balanced accuracy &          Weighted F1 &              ROC AUC \\
\midrule
            CONCH & \textbf{0.643 (0.560, 0.673)} & \textbf{0.600 (0.454, 0.651)} & \textbf{0.873 (0.830, 0.895)} \\
             PLIP & 0.530 (0.520, 0.543) & 0.419 (0.381, 0.461) & 0.701 (0.676, 0.720) \\
       BiomedCLIP & 0.550 (0.513, 0.600) & 0.465 (0.383, 0.577) & 0.726 (0.688, 0.758) \\
       OpenAICLIP & 0.500 (0.500, 0.500) & 0.333 (0.333, 0.333) & 0.564 (0.543, 0.579) \\
\bottomrule
\end{tabular}
\caption{\textbf{Zero-shot classification without prompt ensembling on BRCA subtyping} ($n=150$) in terms of balanced accuracy, weighted F1 score, and ROC AUC. Best performing model for each metric is bolded. 95\% CI is included in parentheses.}
\end{table}

\begin{table}
\centering
\begin{tabular}{llll}
\toprule
       Model name &    Balanced accuracy &          Weighted F1 &              ROC AUC \\
\midrule
            CONCH & \textbf{0.796 (0.756, 0.824)} & \textbf{0.797 (0.748, 0.818)} & \textbf{0.961 (0.956, 0.965)} \\
             PLIP & 0.749 (0.693, 0.764) & 0.741 (0.679, 0.756) & 0.944 (0.939, 0.947) \\
       BiomedCLIP & 0.656 (0.644, 0.682) & 0.639 (0.624, 0.668) & 0.900 (0.890, 0.910) \\
       OpenAICLIP & 0.333 (0.333, 0.338) & 0.167 (0.167, 0.189) & 0.575 (0.562, 0.589) \\
\bottomrule
\end{tabular}
\caption{\textbf{Zero-shot classification without prompt ensembling on RCC subtyping} ($n=225$) in terms of balanced accuracy, weighted F1 score, and ROC AUC (median from 50 sets of randomly sampled prompts). Best performing model for each metric is bolded. 95\% CI is included in parentheses.}
\end{table}

\begin{table}
\centering
\begin{tabular}{llll}
\toprule
       Model name &    Balanced accuracy &          Weighted F1 &              ROC AUC \\
\midrule
            CONCH & \textbf{0.807 (0.720, 0.820)} & \textbf{0.803 (0.705, 0.820)} & \textbf{0.915 (0.889, 0.941)} \\
             PLIP & 0.663 (0.567, 0.700) & 0.640 (0.532, 0.700) & 0.791 (0.763, 0.812) \\
       BiomedCLIP & 0.683 (0.620, 0.727) & 0.670 (0.561, 0.721) & 0.842 (0.805, 0.864) \\
       OpenAICLIP & 0.503 (0.500, 0.517) & 0.371 (0.333, 0.396) & 0.542 (0.513, 0.583) \\
\bottomrule
\end{tabular}
\caption{\textbf{Zero-shot classification without prompt ensembling on NSCLC subtyping} ($n=150$) in terms of balanced accuracy, weighted F1 score, and ROC AUC (median from 50 sets of randomly sampled prompts). Best performing model for each metric is bolded. 95\% CI is included in parentheses.}
\end{table}

\begin{table}
\centering
\begin{tabular}{llll}
\toprule
       Model name &        Cohen's $\kappa$ &    Balanced accuracy &          Weighted F1 \\
\midrule
            CONCH & \textbf{0.116 (0.094, 0.143)} & \textbf{0.324 (0.299, 0.342)} & \textbf{0.244 (0.212, 0.289)} \\
             PLIP & 0.020 (0.014, 0.031) & 0.231 (0.222, 0.246) & 0.114 (0.069, 0.186) \\
       BiomedCLIP & 0.023 (0.014, 0.040) & 0.253 (0.240, 0.273) & 0.067 (0.044, 0.092) \\
       OpenAICLIP & 0.005 (0.000, 0.017) & 0.194 (0.188, 0.205) & 0.119 (0.053, 0.211) \\
\bottomrule
\end{tabular}
\caption{\textbf{Zero-shot classification without prompt ensembling on DHMC LUAD} ($n=143$) in terms of Cohen's $\kappa$, balanced accuracy, and weighted F1 score (median from 50 sets of randomly sampled prompts). Best performing model for each metric is bolded. 95\% CI is included in parentheses.}
\end{table}

\begin{table}
\centering
\begin{tabular}{llll}
\toprule
       Model name & Quadratic weighted $\kappa$ &    Balanced accuracy &          Weighted F1 \\
\midrule
            CONCH &     0.331 (0.254, 0.405) & 0.349 (0.330, 0.390) & 0.245 (0.203, 0.305) \\
             PLIP &     0.163 (0.118, 0.208) & 0.293 (0.267, 0.323) & 0.232 (0.180, 0.264) \\
       BiomedCLIP &    \textbf{ 0.394 (0.355, 0.436)} & \textbf{0.376 (0.351, 0.386)} & \textbf{0.356 (0.319, 0.372)} \\
       OpenAICLIP &     0.003 (0.000, 0.022) & 0.251 (0.250, 0.259) & 0.141 (0.108, 0.200) \\
\bottomrule
\end{tabular}
\caption{\textbf{Zero-shot classification without prompt ensembling on SICAP} ($n=2,122$) in terms of quadratic weighted Cohen's $\kappa$, balanced accuracy, and weighted F1 score (median from 50 sets of randomly sampled prompts). Best performing model for each metric is bolded. 95\% CI is included in parentheses.}
\end{table}

\begin{table}[]
\centering
\begin{tabular}{llll}
\toprule
       Model name &    Balanced accuracy &          Weighted F1 &              ROC AUC \\
\midrule
            CONCH & \textbf{0.566 (0.533, 0.598)} & \textbf{0.542 (0.496, 0.589)} & \textbf{0.901 (0.893, 0.918)} \\
             PLIP & 0.520 (0.485, 0.540) & 0.517 (0.484, 0.576) & 0.879 (0.869, 0.898) \\
       BiomedCLIP & 0.422 (0.398, 0.436) & 0.372 (0.346, 0.408) & 0.859 (0.845, 0.868) \\
       OpenAICLIP & 0.234 (0.211, 0.250) & 0.185 (0.149, 0.202) & 0.727 (0.715, 0.733) \\
\bottomrule
\end{tabular}
\caption{\textbf{Zero-shot classification without prompt ensembling on CRC100k} ($n=7,180$) in terms of balanced accuracy, weighted F1 score, and ROC AUC (median from 50 sets of randomly sampled prompts). Best performing model for each metric is bolded. 95\% CI is included in parentheses.}
\end{table}

\begin{table}
\centering
\begin{tabular}{llll}
\toprule
       Model name &    Balanced accuracy &          Weighted F1 &              ROC AUC \\
\midrule
            CONCH & \textbf{0.598 (0.560, 0.650)} & \textbf{0.590 (0.547, 0.627)} & \textbf{0.795 (0.769, 0.827)} \\
             PLIP & 0.462 (0.439, 0.492) & 0.408 (0.386, 0.475) & 0.710 (0.654, 0.760) \\
       BiomedCLIP & 0.512 (0.474, 0.550) & 0.452 (0.408, 0.518) & 0.747 (0.734, 0.775) \\
       OpenAICLIP & 0.333 (0.328, 0.333) & 0.195 (0.189, 0.239) & 0.551 (0.494, 0.608) \\
\bottomrule
\end{tabular}
\caption{\textbf{Zero-shot classification without prompt ensembling on WSSS4LUAD} ($n=4,693$) in terms of balanced accuracy, weighted F1 score, and ROC AUC (median from 50 sets of randomly sampled prompts). Best performing model for each metric is bolded. 95\% CI is included in parentheses.}
\label{tab:my_label}
\end{table}

\begin{table}
\centering
\begin{tabular}{llll}
\toprule
   Model name &    Balanced accuracy &          Weighted F1 &              ROC AUC \\
\midrule
        CONCH & \textbf{0.847 (0.790, 0.897)} & \textbf{0.844 (0.782, 0.899)} & \textbf{0.937 (0.887, 0.973)} \\
         PLIP & 0.780 (0.713, 0.844) & 0.778 (0.708, 0.846) & 0.864 (0.801, 0.916) \\
   BiomedCLIP & 0.807 (0.747, 0.860) & 0.801 (0.729, 0.858) & 0.911 (0.857, 0.954) \\
   OpenAICLIP & 0.807 (0.745, 0.869) & 0.804 (0.736, 0.867) & 0.899 (0.842, 0.948) \\
ResNet50 (tr) & 0.767 (0.698, 0.834) & 0.765 (0.693, 0.833) & 0.814 (0.740, 0.882) \\
\bottomrule
\end{tabular}
\caption{\textbf{Supervised classification on BRCA subtyping} ($n=150$) in terms of balanced accuracy, weighted F1 score, and ROC AUC. Best performing model for each metric is bolded. 95\% CI is included in parentheses.}
\end{table}

\begin{table}
\centering
\begin{tabular}{llll}
\toprule
   Model name &    Balanced accuracy &          Weighted F1 &              ROC AUC \\
\midrule
        CONCH & \textbf{0.942 (0.910, 0.970)} & \textbf{0.942 (0.910, 0.969)} & \textbf{0.995 (0.989, 0.999)} \\
         PLIP & 0.911 (0.870, 0.947) & 0.911 (0.867, 0.947) & 0.989 (0.979, 0.996) \\
   BiomedCLIP & 0.916 (0.876, 0.948) & 0.915 (0.875, 0.947) & 0.990 (0.980, 0.997) \\
   OpenAICLIP & 0.898 (0.854, 0.935) & 0.898 (0.854, 0.934) & 0.979 (0.957, 0.993) \\
ResNet50 (tr) & 0.893 (0.851, 0.931) & 0.893 (0.852, 0.930) & 0.972 (0.949, 0.989) \\
\bottomrule
\end{tabular}
\caption{\textbf{Supervised classification on RCC subtyping} ($n=225$) in terms of balanced accuracy, weighted F1 score, and ROC AUC. Best performing model for each metric is bolded. 95\% CI is included in parentheses.}
\end{table}

\begin{table}
\centering
\begin{tabular}{llll}
\toprule
   Model name &    Balanced accuracy &          Weighted F1 &              ROC AUC \\
\midrule
        CONCH & \textbf{0.927 (0.886, 0.966)} & \textbf{0.927 (0.886, 0.967)} & \textbf{0.985 (0.968, 0.996)} \\
         PLIP & 0.907 (0.853, 0.952) & 0.907 (0.853, 0.947) & 0.972 (0.944, 0.990) \\
   BiomedCLIP & 0.913 (0.866, 0.959) & 0.913 (0.867, 0.960) & 0.958 (0.923, 0.985) \\
   OpenAICLIP & 0.900 (0.847, 0.946) & 0.900 (0.847, 0.947) & 0.964 (0.935, 0.984) \\
ResNet50 (tr) & 0.840 (0.782, 0.893) & 0.839 (0.780, 0.893) & 0.916 (0.869, 0.956) \\
\bottomrule
\end{tabular}
\caption{\textbf{Supervised classification on NSCLC subtyping} ($n=150$) in terms of balanced accuracy, weighted F1 score, and ROC AUC. Best performing model for each metric is bolded. 95\% CI is included in parentheses.}
\end{table}

\begin{table}
\centering
\begin{tabular}{llll}
\toprule
Model name & Quadratic weighted kappa &    Balanced accuracy &          Weighted F1 \\
\midrule
     CONCH &     \textbf{0.846 (0.828, 0.863)} & \textbf{0.751 (0.729, 0.772)} & \textbf{0.777 (0.759, 0.794)} \\
      PLIP &     0.762 (0.739, 0.784) & 0.589 (0.570, 0.609) & 0.657 (0.637, 0.678) \\
BiomedCLIP &     0.716 (0.694, 0.737) & 0.546 (0.526, 0.565) & 0.625 (0.602, 0.644) \\
OpenAICLIP &     0.704 (0.676, 0.732) & 0.599 (0.579, 0.617) & 0.662 (0.641, 0.682) \\
CTransPath &     0.835 (0.817, 0.851) & 0.678 (0.658, 0.700) & 0.747 (0.728, 0.766) \\
\bottomrule
\end{tabular}
\caption{\textbf{Supervised classification on SICAP} ($n=2,122$) in terms of quadratic weighted Cohen's $\kappa$, balanced accuracy, and weighted F1 score. Best performing model for each metric is bolded. 95\% CI is included in parentheses.}
\label{tab:my_label}
\end{table}

\begin{table}
\centering
\begin{tabular}{llll}
\toprule
Model name &    Balanced accuracy &          Weighted F1 &              ROC AUC \\
\midrule
     CONCH & 0.930 (0.923, 0.937) & 0.949 (0.944, 0.955) & \textbf{0.994 (0.993, 0.995)} \\
      PLIP & 0.879 (0.872, 0.888) & 0.890 (0.884, 0.897) & 0.989 (0.988, 0.991) \\
BiomedCLIP & 0.896 (0.888, 0.903) & 0.921 (0.915, 0.926) & 0.991 (0.989, 0.992) \\
OpenAICLIP & 0.884 (0.877, 0.891) & 0.897 (0.890, 0.904) & 0.992 (0.990, 0.993) \\
CTransPath & \textbf{0.938 (0.932, 0.944)} & \textbf{0.950 (0.945, 0.955)} & \textbf{0.994 (0.993, 0.995)} \\
\bottomrule
\end{tabular}
\caption{\textbf{Supervised classification on CRC100k} ($n=7,180$) in terms of balanced accuracy, weighted F1 score, and ROC AUC. Best performing model for each metric is bolded. 95\% CI is included in parentheses.}
\label{tab:my_label}
\end{table}

\begin{table}
\small 
\centering
\begin{tabular}{lllll}

\toprule
    Model name & Recall@1 & Recall@5 & Recall@10 & Mean Recall\\
\midrule
    CONCH & \textbf{0.400 (0.362, 0.435)} & \textbf{0.776 (0.748, 0.803)} & \textbf{0.888 (0.865, 0.910)} & \textbf{0.688 (0.665, 0.710)} \\
    PLIP & 0.056 (0.042, 0.072) & 0.198 (0.170, 0.226) & 0.307 (0.276, 0.341) & 0.187 (0.165, 0.210) \\
    BiomedCLIP & 0.161 (0.137, 0.189) & 0.405 (0.368, 0.441) & 0.553 (0.517, 0.591) & 0.373 (0.346, 0.402) \\
    OpenAICLIP & 0.018 (0.009, 0.027) & 0.046 (0.032, 0.061) & 0.084 (0.065, 0.103) & 0.049 (0.038, 0.062) \\
\bottomrule
\end{tabular}
\caption{\textbf{Zero-shot text-to-image retrieval performance for Source A} ($n = 797$) in terms of Recall@$K$ for $K\in\{1,5,10\}$ and mean recall over K. Best performing model for each metric is bolded. 95\% CI is included in parentheses.}
\label{tab:source_a_t2i}
\end{table}

\begin{table}
\small 
\centering
\begin{tabular}{lllll}

\toprule
    Model name & Recall@1 & Recall@5 & Recall@10 & Mean Recall\\
\midrule
    CONCH & \textbf{0.402 (0.369, 0.433)} & \textbf{0.792 (0.764, 0.818)} & \textbf{0.883 (0.859, 0.905)} & \textbf{0.692 (0.670, 0.714)} \\
    PLIP & 0.065 (0.049, 0.082) & 0.222 (0.192, 0.251) & 0.320 (0.287, 0.351) & 0.202 (0.179, 0.225) \\
    BiomedCLIP & 0.171 (0.146, 0.197) & 0.439 (0.403, 0.472) & 0.582 (0.547, 0.615) & 0.397 (0.373, 0.423) \\
    OpenAICLIP & 0.010 (0.004, 0.018) & 0.049 (0.035, 0.064) & 0.073 (0.055, 0.092) & 0.044 (0.033, 0.056) \\
\bottomrule
\end{tabular}
\caption{\textbf{Zero-shot image-to-text retrieval performance for Source A} ($n = 797$) in terms of Recall@$K$ for $K\in\{1,5,10\}$ and mean recall over K. Best performing model for each metric is bolded. 95\% CI is included in parentheses.}
\label{tab:source_a_i2t}
\end{table}

\begin{table}
\small 
\centering
\begin{tabular}{lllll}

\toprule
    Model name & Recall@1 & Recall@5 & Recall@10 & Mean Recall\\
\midrule
    CONCH & \textbf{0.171 (0.151, 0.188)} & \textbf{0.437 (0.412, 0.462)} & \textbf{0.562 (0.539, 0.587)} & \textbf{0.390 (0.372, 0.409)} \\
    PLIP & 0.020 (0.013, 0.027) & 0.075 (0.063, 0.089) & 0.132 (0.116, 0.148) & 0.076 (0.066, 0.086) \\
    BiomedCLIP & 0.100 (0.086, 0.115) & 0.264 (0.242, 0.284) & 0.353 (0.328, 0.375) & 0.239 (0.221, 0.256) \\
    OpenAICLIP & 0.009 (0.005, 0.014) & 0.036 (0.027, 0.044) & 0.052 (0.042, 0.063) & 0.032 (0.026, 0.040) \\
\bottomrule
\end{tabular}
\caption{\textbf{Zero-shot text-to-image retrieval performance for Source B} ($n = 1,755$) in terms of Recall@$K$ for $K\in\{1,5,10\}$ and mean recall over K. Best performing model for each metric is bolded. 95\% CI is included in parentheses.}
\label{tab:source_b_t2i}
\end{table}

\begin{table}
\small 
\centering
\begin{tabular}{lllll}

\toprule
    Model name & Recall@1 & Recall@5 & Recall@10 & Mean Recall\\
\midrule
    CONCH & \textbf{0.164 (0.146, 0.181)} & \textbf{0.403 (0.379, 0.426)} & \textbf{0.530 (0.508, 0.553)} & \textbf{0.366 (0.347, 0.384)} \\
    PLIP & 0.024 (0.017, 0.031) & 0.077 (0.065, 0.090) & 0.124 (0.110, 0.140) & 0.075 (0.065, 0.085) \\
    BiomedCLIP & 0.099 (0.085, 0.113) & 0.251 (0.233, 0.272) & 0.359 (0.336, 0.380) & 0.236 (0.220, 0.252) \\
    OpenAICLIP & 0.005 (0.002, 0.008) & 0.027 (0.019, 0.035) & 0.043 (0.034, 0.053) & 0.025 (0.019, 0.031) \\
\bottomrule
\end{tabular}
\caption{\textbf{Zero-shot image-to-text retrieval performance for Source B} ($n = 1,755$) in terms of Recall@$K$ for $K\in\{1,5,10\}$ and mean recall over K. Best performing model for each metric is bolded. 95\% CI is included in parentheses.}
\label{tab:source_b_i2t}
\end{table}

\begin{table}
\small 
\centering
\begin{tabular}{lllll}

\toprule
    Model name & Recall@1 & Recall@5 & Recall@10 & Mean Recall\\
\midrule
    CONCH & \textbf{0.087 (0.047, 0.133)} & \textbf{0.247 (0.180, 0.313)} & \textbf{0.387 (0.313, 0.460)} & \textbf{0.240 (0.189, 0.293)} \\
    PLIP & 0.027 (0.007, 0.053) & 0.080 (0.040, 0.127) & 0.180 (0.120, 0.247) & 0.096 (0.060, 0.136) \\
    BiomedCLIP & 0.053 (0.020, 0.093) & 0.193 (0.133, 0.267) & 0.313 (0.240, 0.387) & 0.187 (0.140, 0.240) \\
    OpenAICLIP & 0.020 (0.000, 0.047) & 0.060 (0.027, 0.100) & 0.107 (0.060, 0.160) & 0.062 (0.033, 0.098) \\
\bottomrule
\end{tabular}
\caption{\textbf{Zero-shot text-to-image retrieval performance for TCGA LUAD} ($n = 165$) in terms of Recall@$K$ for $K\in\{1,5,10\}$ and mean recall over K. Best performing model for each metric is bolded. 95\% CI is included in parentheses.}
\label{tab:mgb_luad_t2i}
\end{table}

\begin{table}
\small 
\centering
\begin{tabular}{lllll}

\toprule
    Model name & Recall@1 & Recall@5 & Recall@10 & Mean Recall\\
\midrule
    CONCH & \textbf{0.067 (0.030, 0.103)} & \textbf{0.188 (0.133, 0.248)} & \textbf{0.291 (0.224, 0.364)} &\textbf{ 0.182 (0.135, 0.230)} \\
    PLIP & 0.018 (0.000, 0.042) & 0.079 (0.042, 0.121) & 0.115 (0.067, 0.164) & 0.071 (0.040, 0.101) \\
    BiomedCLIP & 0.048 (0.018, 0.085) & 0.158 (0.103, 0.212) & 0.291 (0.224, 0.364) & 0.166 (0.125, 0.210) \\
    OpenAICLIP & 0.006 (0.000, 0.018) & 0.042 (0.012, 0.079) & 0.097 (0.055, 0.139) & 0.048 (0.026, 0.075) \\
\bottomrule
\end{tabular}
\caption{\textbf{Zero-shot image-to-text retrieval performance for TCGA LUAD} ($n = 165$) in terms of Recall@$K$ for $K\in\{1,5,10\}$ and mean recall over K. Best performing model for each metric is bolded. 95\% CI is included in parentheses.}
\label{tab:mgb_luad_i2t}
\end{table}

\begin{table}
\centering
\begin{tabular}{llll}
\toprule
Model name &                  Dice score &             Precision &                Recall \\
\midrule
CONCH      &  \textbf{0.601 (0.530, 0.675)} &  \textbf{0.672 (0.630, 0.722)} &  \textbf{0.751 (0.696, 0.803)} \\
PLIP       &  0.549 (0.496, 0.605) &  0.605 (0.556, 0.656) &  0.644 (0.595, 0.694) \\
BiomedCLIP &  0.484 (0.452, 0.520) &  0.536 (0.505, 0.569) &  0.557 (0.519, 0.598) \\
OpenAICLIP &  0.367 (0.314, 0.426) &  0.599 (0.573, 0.629) &  0.605 (0.571, 0.639) \\
\bottomrule
\end{tabular}
\caption{\textbf{Zero-shot segmentation performance on SICAP} ($n=31$) in terms of Dice score as well as precision and recall for the positive class. Best performing model for each metric is bolded. 95\% CI is included in parentheses.}
\label{tab:zeroshot_seg}
\end{table}

\begin{table}
\centering
\begin{tabular}{llll}
\toprule
Model Name &                  Dice &             Precision &                Recall \\
\midrule
CONCH      &  \textbf{0.615 (0.597, 0.633)} &  \textbf{0.663 (0.650, 0.675)} &  \textbf{0.709 (0.693, 0.726)} \\
PLIP       &  0.426 (0.411, 0.443) &  0.526 (0.513, 0.537) &  0.541 (0.528, 0.554) \\
BiomedCLIP &  0.446 (0.430, 0.462) &  0.581 (0.569, 0.592) &  0.601 (0.588, 0.615) \\
OpenAICLIP &  0.367 (0.351, 0.381) &  0.492 (0.481, 0.503) &  0.511 (0.498, 0.524) \\
\bottomrule
\end{tabular}
\caption{\textbf{Zero-shot segmentation performance on DigestPath} ($n=250$) in terms of Dice score as well as precision and recall for the positive class. Best performing model for each metric is bolded. 95\% CI is included in parentheses.}
\end{table}

\begin{table}
\centering
\begin{tabular}{llll}

\toprule
    Model name & METEOR & ROUGE \\
\midrule
    CONCH & \textbf{0.193 (0.181, 0.208)} & \textbf{0.215 (0.200, 0.229)} \\
    GIT-base & 0.122 (0.115, 0.130) & 0.135 (0.125, 0.145) \\
    GIT-large & 0.125 (0.117, 0.134) & 0.153 (0.143, 0.163) \\
\bottomrule
\end{tabular}
\caption{\textbf{Captioning performance with fine-tuning on Source A} (train $n=558$, validation $n=77$, test $n=162$) in terms of METEOR and ROUGE. Best performing model for each metric is bolded. 95\% CI is included in parentheses.}
\label{tab:source_a_captioning}
\end{table}

\begin{table}[h]
  \centering
  \begin{tabular}{@{}p{7.5cm}|p{3cm}}
    \toprule
    Hyperparameter & Value \\
    \midrule
    Automatic mixed precision & fp16 \\
    Batch size & 384 \\
    Gradient accumulation & 4 \\
    Weight decay & 0.2 \\
    AdamW $\beta$ & (0.9, 0.999) \\
    Temperature & Learned \\
    Peak learning rate & 1e-4 \\
    Learning rate schedule & Cosine \\
    Warmup steps & 250 \\
    Epochs & 40 \\
    \bottomrule
  \end{tabular}
  \caption{\textbf{Hyperparameters used in visual-language pretraining}. 8 $\times$ 80GB NVIDIA A100 GPUs were used for training. Effective batch size used for optimization is batch size $\times$ gradient accumulation steps. The maximum sequence length for captions is set to 128.}
  \label{tab:hparams_pretrain}
\end{table}

\begin{table}[h]
  \centering
  \begin{tabular}{p{7.5cm}|p{3cm}}
    \toprule
    Hyperparameter & Value \\
    \midrule
    Layers & 12 \\
    Heads & 12 \\
    Patch size & 16 \\
    Head activation & GELU \\
    Embedding dimension & 768 \\
    Drop path rate & 0.1 \\
    \midrule
    Global crop scale & 0.32, 1.0 \\
    Global crop number & 2 \\
    Local crop scale & 0.05, 0.32 \\
    Local crop number & 10 \\
    Partial prediction shape & Block \\
    Partial prediction ratio & 0.3 \\
    Partial prediction variance & 0.2 \\
    Gradient clipping max norm & 0.3 \\
    Normalize last layer & \checkmark \\
    Shared head & \checkmark \\
    \midrule
    AdamW $\beta$ & (0.9, 0.999) \\
    Batch size & 1024 \\
    Freeze last layer epochs & 3 \\
    Warmup epochs & 10 \\
    Warmup teacher temperature epochs & 30 \\
    Max epochs & 80 \\
    Learning rate schedule & Cosine \\
    Learning rate (start) & 0 \\
    Learning rate (post warmup) & 2e-3 \\
    Learning rate (final) & 2e-6 \\
    Teacher temperature (start) & 0.04 \\
    Teacher temperature (final) & 0.4 \\
    Teacher momentum (start) & 0.996 \\
    Teacher momentum (final) & 1.000 \\
    Weight decay (start) & 0.04 \\
    Weight decay (end) & 0.4 \\
    Automatic mixed precision & fp16 \\
    \bottomrule
  \end{tabular}
  \caption{\textbf{Hyperparameters used in pretraining the vision model}. 4 $\times$ 80GB NVIDIA A100 GPUs were used for training. Batch size refers to the total batch size across GPUs.}
  \label{tab:hparams_ibot}
\end{table}

\begin{table}[h]
  \centering
  \begin{tabular}{p{7.5cm}|p{3cm}}
    \toprule
    Hyperparameter & Value \\
    \midrule
    Layers & 24 \\
    Heads & 12 \\ 
    Embedding dimension & 768 \\
    Hidden dimension & 3,072 \\
    Max. sequence length & 512 \\
    Pos. embedding & Absolute \\
    Vocabulary size & 32,000 \\
    \midrule
    Automatic mixed precision & fp16 \\
    Batch size & 64 \\
    Gradient accumulation & 8 \\
    Weight decay & 0.01 \\
    AdamW $\beta$ & (0.9, 0.999) \\
    Peak learnng rate & 1e-3 \\
    Learning rate schedule & Linear \\
    Warmup steps & 500 \\
    Training steps & 15,000 \\
    \bottomrule
  \end{tabular}
  \caption{\textbf{Hyperparameters used in pretraining the language model}. In-house pathology reports were first de-identified using regex pattern matching before tokenization. 4 $\times$ 80GB NVIDIA A100 GPUs were used for training. Batch size refers to the total batch size across GPUs. Effective batch size used for optimization is batch size $\times$ gradient accumulation steps. The sequence length of training examples was set to the maximum sequence length supported by the model (\textit{i.e.} 512).}
  \label{tab:hparams_gpt}
\end{table}

\begin{table}[h]
  \centering
  \begin{tabular}{@{}p{4cm}|rrr|r@{}}
    \toprule
    Hyperparameter & Value \\
    \midrule
    Batch size & 1 \\
    Weight decay & 1e-5 \\
    AdamW $\beta$ & (0.9, 0.999) \\
    Peak learning rate & 1e-4 \\
    Learning rate schedule & Cosine \\
    Epochs & 20 \\
    \bottomrule
  \end{tabular}
  \caption{\textbf{Hyperparameters used in slide-level supervised classification}. A single 24GB NVIDIA GeForce RTX 3090 GPU was used for each ABMIL model using weakly-supervised learning and slide-level labels.}
  \label{tab:hparams_slide_sup}
\end{table}

\begin{table}[h]
  \centering
  \begin{tabular}{@{}p{4cm}|rrr|r@{}}
    \toprule
    Hyperparameter & Value \\
    \midrule
    Batch size & 16 \\
    Weight decay & 0.2 \\
    AdamW $\beta$ & (0.9, 0.999) \\
    Learning rate & 1e-4 \\
    Warmup steps & 10 \\
    Early stopping patience & 10 \\
    Epochs & 40 \\
    \bottomrule
  \end{tabular}
  \caption{\textbf{Hyperparameters used in caption fine-tuning}. A single 24GB NVIDIA GeForce RTX 3090 GPU was used for training. The maximum sequence length for captions is set to 128. Top-$K$ sampling with $K=50$ was used as decoding strategy at generation time.}
  \label{tab:hparams_supbase}
\end{table}

\begin{table}[]
    \centering
    \begin{tabular}{l}
        \texttt{CLASSNAME.}\\
        \texttt{a photomicrograph showing CLASSNAME.}\\
        \texttt{a photomicrograph of CLASSNAME.}\\
        \texttt{an image of CLASSNAME.}\\
        \texttt{an image showing CLASSNAME.}\\
        \texttt{an example of CLASSNAME.}\\
        \texttt{CLASSNAME is shown.}\\
        \texttt{this is CLASSNAME.}\\
        \texttt{there is CLASSNAME.}\\
        \texttt{a histopathological image showing CLASSNAME.}\\
        \texttt{a histopathological image of CLASSNAME.}\\
        \texttt{a histopathological photograph of CLASSNAME.}\\
        \texttt{a histopathological photograph showing CLASSNAME.}\\
        \texttt{shows CLASSNAME.}\\
        \texttt{presence of CLASSNAME.}\\
        \texttt{CLASSNAME is present.}\\
        \texttt{an H\&E stained image of CLASSNAME.}\\
        \texttt{an H\&E stained image showing CLASSNAME.}\\
        \texttt{an H\&E image showing CLASSNAME.}\\
        \texttt{an H\&E image of CLASSNAME.}\\
        \texttt{CLASSNAME, H\&E stain.}\\
        \texttt{CLASSNAME, H\&E.}\\
    \end{tabular}
    \caption{\textbf{Prompt templates} used for all tasks involving prompts. The name of the class replaces \texttt{CLASSNAME}. See \textbf{Tables 35-38} for class prompts of each task.}
\end{table}

\begin{table*}[]
    \centering
    \begin{tabular}{l|l|l}
        \toprule
        Task & Class & Class names \\
        \midrule
        \multirow{10}{*}{TCGA BRCA} & \multirow{5}{*}{IDC} & \texttt{invasive ductal carcinoma} \\
        & & \texttt{breast invasive ductal carcinoma} \\
        & & \texttt{invasive ductal carcinoma of the breast} \\
        & & \texttt{invasive carcinoma of the breast, ductal pattern} \\
        & & \texttt{breast IDC} \\
        \cmidrule{2-3}
        & \multirow{5}{*}{ILC} & \texttt{invasive lobular carcinoma } \\
        & & \texttt{breast invasive lobular carcinoma} \\
        & & \texttt{invasive lobular carcinoma of the breast} \\
        & & \texttt{invasive carcinoma of the breast, lobular pattern} \\
        & & \texttt{breast ILC} \\
        \midrule
        \multirow{8}{*}{TCGA NSCLC} & \multirow{4}{*}{LUAD} & \texttt{adenocarcinoma} \\
        & & \texttt{lung adenocarcinoma} \\
        & & \texttt{adenocarcinoma of the lung} \\
        & & \texttt{LUAD} \\
        \cmidrule{2-3}
        & \multirow{4}{*}{LUSC} & \texttt{squamous cell carcinoma} \\
        & & \texttt{lung squamous cell carcinoma} \\
        & & \texttt{squamous cell carcinoma of the lung} \\
        & & \texttt{LUSC} \\
        \midrule
        \multirow{12}{*}{TCGA RCC} & \multirow{4}{*}{CCRCC} & \texttt{clear cell renal cell carcinoma} \\
        & & \texttt{renal cell carcinoma, clear cell type} \\
        & & \texttt{renal cell carcinoma of the clear cell type} \\
        & & \texttt{clear cell RCC} \\
        \cmidrule{2-3}
        & \multirow{4}{*}{PRCC} & \texttt{papillary renal cell carcinoma} \\
        & & \texttt{renal cell carcinoma, papillary type} \\
        & & \texttt{renal cell carcinoma of the papillary type} \\
        & & \texttt{papillary RCC} \\
        \cmidrule{2-3}
        & \multirow{4}{*}{CHRCC} & \texttt{chromophobe renal cell carcinoma} \\
        & & \texttt{renal cell carcinoma, chromophobe type} \\
        & & \texttt{renal cell carcinoma of the chromophobe type} \\
        & & \texttt{chromophobe RCC} \\
    \bottomrule
    \end{tabular}
    \caption{\textbf{Class prompts for BRCA, NSCLC, and RCC subtyping}.}
\end{table*}

\begin{table*}[]
    \centering
    \footnotesize
    \begin{tabular}{p{1cm}|l|l}
        \toprule
        Task & Class & Class names \\
        \midrule
        \multirow{25}{*}{\shortstack{DHMC\\LUAD}} & \multirow{5}{*}{papillary} & \texttt{papillary pattern adenocarcinoma} \\
        & & \texttt{papillary pattern adenocarcinoma of the lung} \\
        & & \texttt{lung adenocarcinoma, papillary growth pattern} \\
        & & \texttt{lung adenocarcinoma with a predominantly papillary growth pattern} \\
        & & \texttt{lung adenocarcinoma, papillary predominant histological subtype} \\
        \cmidrule{2-3}
        & \multirow{5}{*}{solid} & \texttt{solid pattern adenocarcinoma} \\
        & & \texttt{solid pattern adenocarcinoma of the lung} \\
        & & \texttt{lung adenocarcinoma, solid growth pattern} \\
        & & \texttt{lung adenocarcinoma with a predominantly solid growth pattern} \\
        & & \texttt{lung adenocarcinoma, solid predominant histological subtype} \\
        \cmidrule{2-3}
        & \multirow{5}{*}{micropapillary} & \texttt{micropapillary pattern adenocarcinoma} \\
        & & \texttt{micropapillary pattern adenocarcinoma of the lung} \\
        & & \texttt{lung adenocarcinoma, micropapillary growth pattern} \\
        & & \texttt{lung adenocarcinoma with a predominantly micropapillary growth pattern} \\
        & & \texttt{lung adenocarcinoma, micropapillary predominant histological subtype} \\
        \cmidrule{2-3}
        & \multirow{5}{*}{acinar} & \texttt{acinar pattern adenocarcinoma} \\
        & & \texttt{acinar pattern adenocarcinoma of the lung} \\
        & & \texttt{lung adenocarcinoma, acinar growth pattern} \\
        & & \texttt{lung adenocarcinoma with a predominantly acinar growth pattern} \\
        & & \texttt{lung adenocarcinoma, acinar predominant histological subtype} \\
        \cmidrule{2-3}
        & \multirow{5}{*}{leipidic} & \texttt{leipidic pattern adenocarcinoma} \\
        & & \texttt{leipidic pattern adenocarcinoma of the lung} \\
        & & \texttt{lung adenocarcinoma, leipidic growth pattern} \\
        & & \texttt{lung adenocarcinoma with a predominantly leipidic growth pattern} \\
        & & \texttt{lung adenocarcinoma, leipidic predominant histological subtype} \\
    \bottomrule
    \end{tabular}
    \caption{\textbf{Class prompts for DHMC LUAD pattern classification}.}
\end{table*}

\begin{table*}[]
    \centering
    \begin{tabular}{l|l|l}
        \toprule
        Task & Class & Class names \\
        \midrule
        \multirow{41}{*}{CRC100k} & \multirow{5}{*}{ADI} & \texttt{adipose} \\
        & & \texttt{adipose tissue} \\
        & & \texttt{adipocytes} \\
        & & \texttt{fat} \\
        & & \texttt{fat cells} \\
        \cmidrule{2-3}
        & \multirow{4}{*}{BACK} & \texttt{background} \\
        & & \texttt{penmarking} \\
        & & \texttt{empty space} \\
        & & \texttt{background artifacts} \\
        \cmidrule{2-3}
        & \multirow{4}{*}{DEB} & \texttt{debris} \\
        & & \texttt{colorectal adenocarcinoma debris and necrosis} \\
        & & \texttt{necrosis} \\
        & & \texttt{necrotic debris} \\
        \cmidrule{2-3}
        & \multirow{5}{*}{LYM} & \texttt{lymphocytes} \\
        & & \texttt{lymphoid aggregate} \\
        & & \texttt{immune cells} \\
        & & \texttt{lymphoid infiltrate} \\
        & & \texttt{inflammatory cells} \\
        \cmidrule{2-3}
        & \multirow{4}{*}{MUC} & \texttt{mucus} \\
        & & \texttt{mucin} \\
        & & \texttt{mucus pool} \\
        & & \texttt{mucin pool} \\
        \cmidrule{2-3}
        & \multirow{5}{*}{MUS} & \texttt{smooth muscle} \\
        & & \texttt{smooth muscle tissue} \\
        & & \texttt{muscle} \\
        & & \texttt{muscularis propria} \\
        & & \texttt{muscularis mucosa} \\
        \cmidrule{2-3}
        & \multirow{4}{*}{NORM} & \texttt{normal colon mucosa} \\
        & & \texttt{uninvolved colon mucosa} \\
        & & \texttt{benign colon mucosa} \\
        & & \texttt{benign epithelium} \\
        \cmidrule{2-3}
        & \multirow{5}{*}{STR} & \texttt{cancer-associated stroma} \\
        & & \texttt{tumor-associated stroma} \\
        & & \texttt{stromal cells} \\
        & & \texttt{stromal tissue} \\
        & & \texttt{stroma} \\
        \cmidrule{2-3}
        & \multirow{5}{*}{TUM} & \texttt{colorectal adenocarcinoma epithelium} \\
        & & \texttt{colorectal adenocarcinoma} \\
        & & \texttt{tumor} \\
        & & \texttt{adenocarcinoma} \\
        & & \texttt{malignant epithelium} \\
    \bottomrule
    \end{tabular}
    \caption{\textbf{Class prompts for CRC100k}. The Class column refers to the original class names used in the CRC100k dataset\cite{kather2019predicting}.}
\end{table*}

\begin{table*}[]
    \centering
    \footnotesize
    \begin{tabular}{l|l|l}
        \toprule
        Task & Class & Class names \\
        \midrule
        \multirow{10}{*}{WSSS4LUAD} & \multirow{3}{*}{normal} & \texttt{non-tumor} \\
        & & \texttt{normal tissue} \\
        & & \texttt{non-cancerous tissue} \\
        \cmidrule{2-3}
        & \multirow{4}{*}{stroma} & \texttt{tumor-associated stroma} \\
        & & \texttt{cancer-associated stroma} \\
        & & \texttt{tumor-associated stromal tissue} \\
        & & \texttt{cancer-associated stromal tissue} \\
        \cmidrule{2-3}
        & \multirow{3}{*}{tumor} & \texttt{tumor tissue} \\
        & & \texttt{tumor epithelial tissue} \\
        & & \texttt{cancerous tissue} \\
        \midrule
        \multirow{29}{*}{SICAP} & \multirow{6}{*}{NC} & \texttt{non-cancerous tissue} \\
        & & \texttt{non-cancerous prostate tissue} \\
        & & \texttt{benign tissue} \\
        & & \texttt{benign glands} \\
        & & \texttt{benign prostate tissue} \\
        & & \texttt{benign prostate glands} \\
        \cmidrule{2-3}
        & \multirow{6}{*}{G3} & \texttt{gleason grade 3} \\
        & & \texttt{gleason pattern 3} \\
        & & \texttt{prostate cancer, gleason grade 3} \\
        & & \texttt{prostate cancer, gleason pattern 3} \\
        & & \texttt{prostate adenocarcinoma, well-differentiated} \\
        & & \texttt{well-differentiated prostatic adenocarcinoma} \\
        \cmidrule{2-3}
        & \multirow{6}{*}{G4} & \texttt{gleason grade 4} \\
        & & \texttt{gleason pattern 4} \\
        & & \texttt{prostate cancer, gleason grade 4} \\
        & & \texttt{prostate cancer, gleason pattern 4} \\
        & & \texttt{prostate adenocarcinoma, moderately differentiated} \\
        & & \texttt{moderately differentiated prostatic adenocarcinoma} \\
        \cmidrule{2-3}
        & \multirow{6}{*}{G5} & \texttt{gleason grade 5} \\
        & & \texttt{gleason pattern 5} \\
        & & \texttt{prostate cancer, gleason grade 5} \\
        & & \texttt{prostate cancer, gleason pattern 5} \\
        & & \texttt{prostate adenocarcinoma, poorly differentiated} \\
        & & \texttt{poorly differentiated prostatic adenocarcinoma} \\
        \cmidrule{2-3}
        & \multirow{5}{*}{Tumor} & \texttt{prostatic adenocarcinoma} \\
        & & \texttt{adenocarcinoma} \\
        & & \texttt{prostate cancer} \\
        & & \texttt{tumor tissue} \\
        & & \texttt{cancerous tissue} \\
        \midrule
        \multirow{9}{*}{DigestPath} & \multirow{4}{*}{Benign} & \texttt{benign tissue} \\
        & & \texttt{benign colon tissue} \\
        & & \texttt{benign colorectal tissue} \\
        & & \texttt{benign rectal tissue} \\
        \cmidrule{2-3}
        & \multirow{4}{*}{Malignant} & \texttt{malignant tissue} \\
        & & \texttt{malignant colon tissue} \\
        & & \texttt{malignant colorectal tissue} \\
        & & \texttt{malignant rectal tissue} \\
    \bottomrule
    \end{tabular}
    \caption{\textbf{Class prompts for WSSS4LUAD, SICAP, and DigestPath}. SICAP ROI-level classification uses the prompts for NC through G5. SICAP slide-level tumor segmentation uses prompts for NC and Tumor.}
\end{table*}

\clearpage

\begin{nolinenumbers}
\section*{References} 
\vspace{2mm}

\begin{spacing}{0.9}
\bibliographystyle{naturemag}
\bibliography{sample}
\end{spacing}
\end{nolinenumbers}

\end{document}